\documentclass{article}

\PassOptionsToPackage{numbers, compress}{natbib}
\usepackage[preprint]{neurips_2026}
\usepackage[utf8]{inputenc} 
\usepackage[T1]{fontenc}    
\usepackage{fonttable}
\usepackage[english]{babel} 
\usepackage[protrusion=true,expansion=true]{microtype}
\usepackage{lmodern}
\usepackage{comment}
\usepackage{cancel}
\usepackage{csquotes}
\usepackage{listings}
\usepackage{nameref}
\usepackage{paralist}
\usepackage{xspace}
\usepackage{setspace}
\usepackage{makeidx}
\usepackage{pdfpages}
\usepackage{times}

\usepackage{soul}

\usepackage{amsmath,amssymb} 
\usepackage{amsfonts}       
\usepackage{mathrsfs} 
\usepackage{mathtools}
\usepackage{array} 
\usepackage{nicefrac}       
\usepackage{textcomp}
\usepackage{bm} 
\usepackage{pifont} 

\usepackage{graphicx}
\usepackage{wrapfig}
\usepackage{adjustbox} 
\usepackage{epsfig}
\pdfoutput=1
\graphicspath{{./figures/}}
\DeclareGraphicsExtensions{.pdf,.jpeg,.png,.jpg}

\usepackage{float}
\usepackage{stfloats}
\usepackage{placeins}
\usepackage{subcaption}
\usepackage[font=small,labelfont=bf]{caption}
\usepackage{lscape}      

\usepackage{tikz}  
\usepackage{makecell}
\usepackage{overpic}
\usepackage{cuted} 

\newsavebox{\subfigbox}

\usepackage{algorithm}
\usepackage[noend]{algpseudocode}

\usepackage{array} 
\usepackage{tabularx}
\usepackage{multirow}
\usepackage{multicol}
\usepackage{booktabs}   
\usepackage{tablefootnote}
\usepackage{threeparttable}

\usepackage{paralist}
\usepackage{enumitem}
\setlist[itemize]{noitemsep,leftmargin=*,topsep=0in}
\setlist[enumerate]{noitemsep,leftmargin=*,topsep=0in}

\usepackage{url}            
\usepackage[table]{xcolor}
\usepackage{color,colortbl} 
\usepackage{soul}

\PassOptionsToPackage{capitalize, noabbrev}{cleveref}

\usepackage{hyperref}

\usepackage{titlesec}
\titlespacing{\section}{0pt}{0.15\baselineskip}{0.1\baselineskip}
\titlespacing{\subsection}{0pt}{0.1\baselineskip}{0.08\baselineskip}
\titlespacing{\subsubsection}{0pt}{0.03\baselineskip}{0.02\baselineskip}

\renewcommand{\paragraph}[1]{\noindent\textbf{#1}: }

\usepackage{blindtext}
\usepackage{bbm}

\usepackage{lipsum}

\definecolor{color1}{rgb}{0.6, 0.4, 0.05}
\definecolor{color2}{rgb}{0.0, 0.7, 0.7}
\definecolor{color3}{rgb}{0.35, 0.75, 0.0}
\definecolor{color4}{rgb}{0.4, 0.8, 0.0}
\definecolor{color5}{rgb}{0.5, 0.0, 0.5}
\definecolor{color6}{rgb}{0.99, 0.84, 0.69}
\definecolor{color7}{rgb}{0.8, 0.3, 0.3}
\definecolor{color8}{rgb}{0.2, 0.4, 0.8}
\definecolor{color9}{rgb}{0.6, 0.2, 0.6}
\definecolor{revision_color}{rgb}{1,0,0}

\hypersetup{
    colorlinks=true,
    linkcolor=black,   
    citecolor=black,   
    urlcolor=blue      
}

\usepackage{tikz}
\usetikzlibrary{arrows.meta, positioning, fit, backgrounds, shapes.geometric}

\definecolor{orangebox}{RGB}{245,200,150}
\definecolor{greenbox}{RGB}{180,220,170}
\definecolor{purplebox}{RGB}{200,190,230}
\definecolor{graybox}{RGB}{220,220,220}
\definecolor{lightblue}{RGB}{210,230,245}
\definecolor{attentionblue}{RGB}{100,160,210}
\definecolor{decoderblue}{RGB}{180,210,235}
\definecolor{labelgray}{RGB}{100,100,100}

\title{Less Decoder is More Encoder: Geometric Representation Learning from Novel View Synthesis}

\author{
  \textbf{
    Keerthi Kaashyap \quad
    Dennis Anthony \quad
    Akshay Krishnan \quad
    Nhi Ngoc Nguyen \quad
  }\\[0.15cm]
  \textbf{
   Jeremy Collins \quad
    James Hays \quad
    Shreyas Kousik \quad
    Animesh Garg
  }\\[0.20cm]
  Georgia Institute of Technology
}

\begin{document}
\titlespacing{\section}{0pt}{0.3\baselineskip}{0.25\baselineskip}
\titlespacing{\subsection}{0pt}{0.2\baselineskip}{0.15\baselineskip}
\titlespacing{\subsubsection}{0pt}{0.05\baselineskip}{0.03\baselineskip}

\renewcommand{\paragraph}[1]{\noindent\textbf{#1} --}

\newcommand{\method}{\textsc{SNAP}\xspace} 
\newcommand{\prope}{\textsc{PRoPE}\xspace}
\newcommand{\dinovthree}{DINOv3\xspace}
\newcommand{\blocknn}{\text{BlockNN}\xspace}

\newcommand{\encoder}{\mathcal{E}}
\newcommand{\decoder}{\mathcal{D}}
\newcommand{\policy}{\pi}

\newcommand{\numviews}{N}
\newcommand{\numpatches}{L}
\newcommand{\embdim}{d}
\newcommand{\image}{\mathbf{I}}
\newcommand{\targetimage}{\image_t}
\newcommand{\pose}{\mathbf{P}}
\newcommand{\targetpose}{\pose_t}
\newcommand{\imgfeatures}{\mathbf{F}}
\newcommand{\targetfeatures}{\hat{\imgfeatures}}
\newcommand{\scenetokens}{\mathbf{Z}}
\newcommand{\latent}{\mathbf{z}} 
\newcommand{\queries}{\mathbf{Q}}
\newcommand{\ci}[1]{\,{\scriptsize$\pm$#1}}

\newcommand{\rf}{\text{RF}\xspace}
\newcommand{\lrecon}{\mathcal{L}_{\text{recon}}}
\newcommand{\separability}{\mathcal{S}}

\newcommand{\kk}[1]{{\color{green} [KK: #1]}}
\newcommand{\ak}[1]{{\color{red} [AK: #1]}}
\newcommand{\shrey}[1]{{\color{red} [SK: #1]}}
\newcommand{\den}[1]{{\color{blue} [DA: #1]}} 

\maketitle

\begin{abstract}
  This paper examines the role of Novel View Synthesis (NVS) in geometric representation learning. 
  In principle, NVS should reason about 3D scene structure, thereby enabling transferable multi-view geometric representations. Yet, existing encoder-based NVS methods yield poor representations. This is not because of a lack of supervisory signal, but rather due to inconspicuous architectural choices: \textit{spatially expressive decoders} that dilute representational capabilities of the scene encoder, and \textit{low-level pixel-space targets} that hinder feature learning. 
  We present \method, a self-supervised encoder-decoder transformer that addresses both through a pose-conditioned local decoder and a latent-space reconstruction objective.
  \method is task agnostic, and we show that it is competitive with special-purpose geometry-supervised methods.
  \method also performs competitively against self-supervised representations across five tasks: visual localization, pose estimation, point correspondence, depth estimation, and robot manipulation.
  Remarkably, \method's patch features exhibit emergent viewpoint invariance that approaches heavily supervised models despite lower compute and data budgets. Under camera shifts where standard 2D representations collapse, \method\ degrades more gracefully, revealing that restricting decoder expressivity actively prevents the suppression of transferable geometric structure. \url{https://snap-nvs.github.io}
\end{abstract}

\section{Introduction}

Geometric understanding from images is a foundation of computer vision. 
Methods like DUSt3R~\cite{wang2024dust3r}, MASt3R~\cite{leroy2024mast3r}, and VGGT~\cite{wang2025vggt} demonstrate that transformers can produce rich, transferable geometric representations, when trained on large-scale annotated 3D data.
This supervision requirement is a fundamental bottleneck since dense 3D ground truth is expensive to collect at scale.

Novel View Synthesis (NVS) offers a compelling alternative \cite{jin2025lvsm,zhao2026erayzer,jiang2025rayzer,kim2026svsm}.
These models must reason about 3D scene structure to predict unseen viewpoints, providing a rich geometric supervision signal only from posed images, which are abundantly available across videos, robotics platforms, and structure-from-motion pipelines.
However, representations produced from feed-forward NVS methods transfer poorly to downstream geometric tasks.
We argue this poor transfer is not a signal failure, but rather an architectural failure, driven by two identifiable and independently addressable choices that dilute signal.

\textbf{First}, spatially expressive decoders tend to absorb the representational load of the task.
When a decoder has sufficient capacity to reconstruct a target view from any encoder output, the encoder is not pressured to maintain structured, locally discriminative features.
While the reconstruction task is solved, the latent representations are not geometrically grounded.
\textbf{Second}, pixel-space reconstruction targets require modeling high-frequency aliasing statistics that are weakly related to 3D structure and semantics. Consequently, the reconstruction loss produces gradients that encourage the representation encoder to preserve fine-grained, image-specific statistics that are not directly informative about the underlying 3D structure or semantics.

We present \method, which addresses both failure axes with two targeted interventions. First, a \textit{pose-conditioned local decoder} built on \prope~\cite{li2025cameras} restricts the decoder's receptive field via an explicit \blocknn mask applied to queried-view self-attention. This constraint forces the geometric structure to reside within the encoder's scene representation rather than being an effect of the generative capacity of the decoder. Second, \method operates directly on frozen \dinovthree ViT-B patch features to provide a richer supervision signal than RGB reconstruction. This structured grounding is jointly aware of appearance, geometry, and semantics and produces higher quality representations. 

Our main contributions are:
\begin{enumerate}
  \item \textbf{\method}, a self-supervised multi-view spatial reasoning backbone that builds upon frozen 2D semantic latent spaces with an asymmetric NVS prediction objective, and latent feature targets to learn geometrically transferable representations without dense 3D supervision.
  \item \textbf{Mechanistic analyses} isolating the independent impact of decoder expressivity and prediction space on geometric transfer, via receptive field sweeps and prediction space landscape experiments.
  \item \textbf{Empirical evaluation} across \textit{five tasks} matching or exceeding self-supervised baselines and competitive with geometry-supervised methods that require large-scale annotated 3D data.
\end{enumerate}

\section{Related Work}

\method\ sits at the intersection of self-supervised representation learning, novel view synthesis, and geometric foundation models. Detailed discussions are provided in Appendix~\ref{sec:appendix_related_work}.

\textbf{Self-Supervised Representation Learning.}
Masked Image Modeling (MIM) has evolved from pixel-reconstruction (MAE~\cite{he2022mae}, SimMIM~\cite{xie2022simmim}) toward latent-space prediction (I-JEPA~\cite{assran2023ijepa}, V-JEPA~\cite{bardes2024vjepa}). In multi-view regimes, CroCo~\cite{weinzaepfel2022croco} and Muskie~\cite{li2025muskiemultiviewmaskedimage} leverage cross-view completion to learn implicit 3D priors. \method\ extends this trajectory by utilizing frozen \dinovthree~\cite{simeoni2025dinov3} features as opposed to pixel space targets used in Muskie~\cite{li2025muskiemultiviewmaskedimage}, which act as semantically structured latent distillation targets to provide a cleaner supervision signal for optimization.

\textbf{Novel View Synthesis (NVS).}
Feed-forward NVS was established by architectures such as SRT~\cite{sajjadi2022srt} and RUST~\cite{sajjadi2023rust}, recently scaling to large-view models such as LVSM~\cite{jin2025lvsm}, RayZer~\cite{jiang2025rayzer}, with contemporaneous works LagerNVS~\cite{szymanowicz2026lagernvs} and SVSM~\cite{kim2026svsm} continuing to extend the paradigm. However, most of these variants utilize expressive decoders that absorb the representational load of the task, leading to ungrounded latents. By constraining the decoder's receptive field via a \blocknn mask, adopting \prope~\cite{li2025cameras}, and targeting a latent feature space, \method\ ensures its representations prioritize 3D structural awareness over reconstruction statistics.

\textbf{Geometry-Supervised Models.}
Recent foundation models, including DUSt3R~\cite{wang2024dust3r}, MASt3R~\cite{leroy2024mast3r}, VGGT~\cite{wang2025vggt}, and D4RT~\cite{zhang2026d4rt}, achieve state-of-the-art performance but rely on massive 3D-annotated datasets. Similarly, $\pi^3$~\cite{wang2026pi3} and Depth Anything 3~\cite{lin2026depth} provide robust spatial priors via explicit 3D supervision. Conversely, \method\ approaches the geometric transferability of these models while remaining self-supervised and training on significantly less data and compute.

\section{Method}

\begin{figure}[!t]
  \centering
  \includegraphics[width=0.99\linewidth]{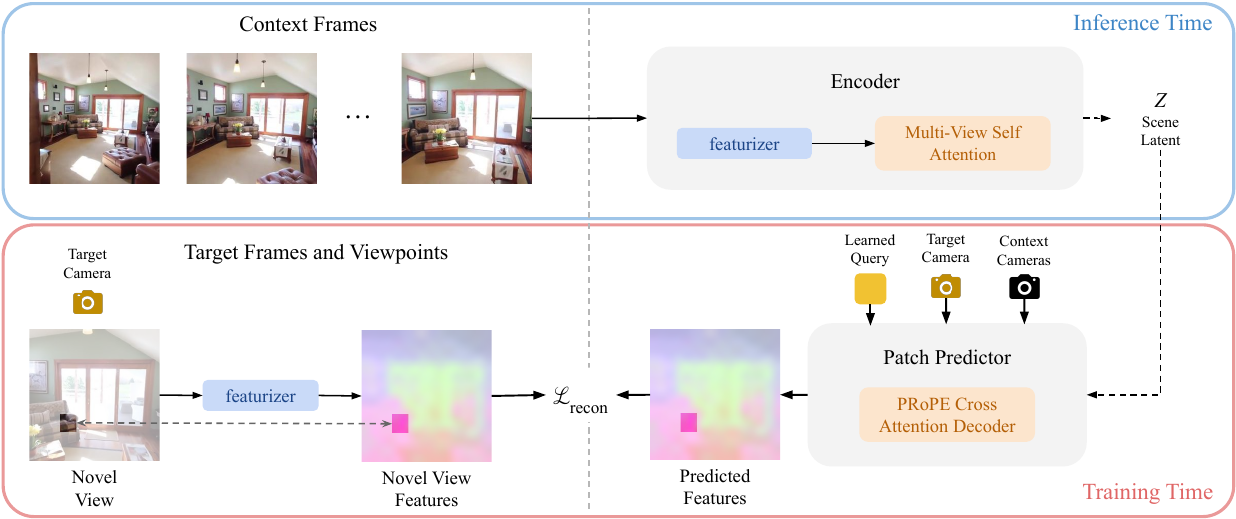}
  \caption{
     \textbf{Training and inference pathways.} \textbf{Top (Inference):} A pose-free multi-view encoder processes reference frames via a frozen \dinovthree\ featurizer ($\varphi$) and cross-view self-attention to construct an implicit 3D scene latent $Z$. \textbf{Bottom (Training):} A pose-conditioned local decoder queries $Z$ using \prope\ to predict the feature map of a novel target view. Supervision ($\mathcal{L}_{\mathrm{recon}}$) is applied in a semantic latent space against a frozen \dinovthree\ teacher, completely bypassing pixel-level rendering and forcing geometric structure into the encoder.
  }
  \label{fig:architecture}
\end{figure}

\method\ operates in two stages, as illustrated in Figure~\ref{fig:architecture}. First, a pose-free multi-view encoder aggregates frozen \dinovthree\ patch features from $\numviews$ reference views into scene tokens $\scenetokens$ via full cross-view self-attention. 
Notably, no camera parameters are provided at this stage. Second, a pose-conditioned patch decoder then queries $\scenetokens$ using \prope-rotated cross-attention to synthesize the target view's feature map. The target and context  camera intrinsics and extrinsics are fed into \prope at this stage. 
Two design choices jointly force geometric structure into the encoder: a \blocknn\ mask restricting the decoder's spatial receptive field (Sec.~\ref{sec:prope_decoder}), and a latent reconstruction objective against frozen \dinovthree\ teacher features rather than RGB pixels (Sec.~\ref{sec:objective}). 
We train \method on scenes from RealEstate10K \cite{zhou2018stereo}, DL3DV \cite{ling2024dl3dv}, and Co3Dv2 \cite{reizenstein2021co3d}; comprehensive implementation and architectural details are deferred to Appendix \ref{sec:appendix_methods}

\subsection{Notation}
\label{sec:formulation}

We use $\{\image_k, T_k, \mathbf{K}_k\}_{k=1}^{\numviews}$ to denote $\numviews$ reference views, where $\image_k \in \mathbb{R}^{3 \times H \times W}$ is an RGB image, $T_k \in \mathrm{SE}(3)$ its camera-to-world pose with respect to $T_1$ (with $T_1$ set as the identity) and
$\mathbf{K}_k \in \mathbb{R}^{3 \times 3}$ its intrinsic matrix.
During inference, a set of target poses $\{T^*_j\}$ and intrinsics 
$\{\mathbf{K}^*_j\}$ are provided for one or more target views. $\numpatches = (H/p) \times (W/p)$ denotes the number of $p{\times}p$ patches per image, $\varphi$ a frozen ViT featurizer (\dinovthree~\cite{simeoni2025dinov3}) with output dimension $\embdim$, and the projection matrix $\pose_k = \hat{\mathbf{K}}_k \cdot T_k^{-1}$ maps world-space 3D coordinates to image-space patch coordinates with normalized intrinsics $\hat{\mathbf{K}}_k$.

\subsection{Pose-Free Multi-View Encoder as a Spatial Backbone}
\label{sec:encoder}
\method's pose-free multi-view encoder aggregates reference features $\imgfeatures_k = \varphi(\image_k) \in \mathbb{R}^{\numpatches \times \embdim}$ into unified scene tokens $\scenetokens = \encoder_{\mathrm{RoPE2D}}\!\bigl(\{F_k\}\bigr) \in \mathbb{R}^{\numviews\numpatches \times E}$.
Each reference image $\image_k$ is patchified by the frozen \dinovthree\ encoder into $\imgfeatures_k \in \mathbb{R}^{\numpatches \times \embdim}$, then projected to the transformer working dimension $E$ via a learned linear layer to yield $X_k \in \mathbb{R}^{\numpatches \times E}$.
RoPE2D~\cite{heo2024rotary} is applied, to queries and keys within self-attention layers, encoding the intra-image 2D patch grid position $(i,j)$.
The projected multi-view sequence $[X_1;\,\ldots;\,X_{\numviews}] \in \mathbb{R}^{\numviews\numpatches \times E}$ is processed by a multi-view transformer with $L_e$ cross-view self-attention blocks.
In full cross-view self-attention, each patch attends to all patches across all reference views. 
The output $\scenetokens$ is a pose-free scene representation with implicit geometry.

\subsection{Pose-Conditioned Patch Decoder as a Feature Predictor}
\label{sec:prope_decoder}

For the second stage of \method, a pose-conditioned patch decoder predicts the target feature map:
\begin{equation}
  \targetfeatures = \decoder_{\prope}\!\bigl(\scenetokens;\,\{\pose^*_j\},\{\pose_k\}\bigr) \in \mathbb{R}^{\numpatches \times \embdim}.
\end{equation}
Our decoder $\decoder$ is a latent novel-view predictor, which predicts the patchified latent features $\imgfeatures^*$ from the camera parameters of the target and context views.
We employ \prope \cite{li2025cameras} for pose-conditioning in the decoder, which consists of alternating blocks of \prope-rotated cross-attention against $\scenetokens$ and block-masked self-attention among target queries which is motivated in Sec.~\ref{sec:axis1}.
A single learned feature query $\mathbf{q} \in \mathbb{R}^{1 \times E}$ is broadcast into a sequence of $\numpatches$ identical target patch queries $\queries \in \mathbb{R}^{\numpatches \times E}$. Spatial symmetry is broken entirely by \prope, which differentiates the learned queries across viewpoints using the target intrinsics $\mathbf{K}^*_j$ and pose $T^*_j$.
During cross-attention, the target queries $\queries$ serve as $q$, while the encoder's scene tokens $\scenetokens$ provide $k$ and $v$. 
\prope constructs projection matrices $\pose_k = \hat{\mathbf{K}}_k \cdot T_k^{-1}$ and rotates queries and keys such that:
\begin{equation}
  \mathrm{Attn}(q,\,k,\,v)
  = \mathrm{softmax}\!\left(
      \frac{(\pose^*_j{}^{\!\top} q)\,(\pose_{k}^{-1} k)^{\!\top}}{\sqrt{E/H}}
    \right)\!v,
  \label{eq:prope-attn}
\end{equation}

where $H$ is the number of attention heads and $\pose^*_j = \hat{\mathbf{K}}^*_j \cdot T^{*-1}_j$ is the projection matrix for target view $j$.
A token pair scores high attention when their world-space rays are co-aligned emulating a soft epipolar search.

Following cross-attention, lateral communication among target queries is restricted by a non-overlapping \blocknn\ mask. Every patch is assigned to an exclusive spatial block of size $\rf{\times}\rf$. Self-attention is only permitted between queries falling within the same block, explicitly bottlenecking the spatial receptive field of the decoder. As demonstrated in Sec.~\ref{sec:axis1}, isolating spatial communication forces geometric reasoning into the encoder; therefore, in our optimal \method\ architecture, we strictly enforce $\rf=1$, reducing target-query self-attention to a pointwise residual MLP.

\subsection{Latent Reconstruction Objective}
\label{sec:objective}

Training minimizes
\begin{equation}
  \lrecon
  = \underbrace{
      \frac{1}{\numpatches\embdim}
      \sum_{p,d}
        \bigl|\targetfeatures_{p,d} - \imgfeatures^*_{p,d}\bigr|
    }_{\mathcal{L}_{\mathrm{feat}}}
  + \frac{1}{2}
    \underbrace{
      \vphantom{\frac{1}{\numpatches\embdim}\sum_{p,d}}
      \bigl(
        \|\nabla_x\targetfeatures - \nabla_x\imgfeatures^*\|_1
       +\|\nabla_y\targetfeatures - \nabla_y\imgfeatures^*\|_1
      \bigr)
    }_{\mathcal{L}_{\mathrm{grad}}},
  \label{eq:objective}
\end{equation}
where $\nabla_x, \nabla_y$ are first-order spatial finite-differences over the 2D token grid. We apply an $L_1$ penalty rather than cosine similarity to the normalized features to promote sharper, sparsity-inducing gradients at structural boundaries.
Both $\targetfeatures$ and the ground-truth target map
$\imgfeatures^* = \varphi(\image^*)$ are $\ell_2$-normalised onto the unit
hypersphere before supervision.

$\mathcal{L}_{\mathrm{grad}}$ is not auxiliary: without it, $\decoder$ can
satisfy $\mathcal{L}_{\mathrm{feat}}$ with spatially-uniform predictions that
average out local structure, collapsing the geometric signal the encoder is
required to encode.
Penalising finite-difference residuals enforces spatial fidelity precisely at
object boundaries and texture edges, where patch-wise $\ell_1$ is least
discriminative.

By routing supervision through \dinovthree features rather than RGB, we replace a weak, low-level training signal with more structured guidance. Unlike raw pixels, which contain nuisance variation and little inherent bias toward semantics, \dinovthree features reside on a structured manifold and provide dense, spatially localized embeddings. Section \ref{sec:axis2} motivates this choice and quantifies its effect via a systematic sweep over \dinovthree layers.

\subsection{Multitask Evaluation of Representation}
\label{sec:downstream}

We probe the geometric content of frozen $\scenetokens$ on five tasks spanning
correspondence, localization, and pose estimation.
In every case $\encoder$ is held frozen; only a lightweight task-specific head
is trained.

\paragraph{(T1) Point Correspondence} Given views $A$ and $B$, a point correspondence head
\texttt{PC}$(\scenetokens_A,\,\scenetokens_B,\,\mathbf{c}_A)$
predicts the normalized coordinate $\hat{\mathbf{c}}_B \in [0,1]^2$ in view
$B$ matching query $\mathbf{c}_A \in [0,1]^2$ in view $A$.
Both views are encoded under identity poses; pixel-unit intrinsics
$(f_x, f_y, c_x, c_y)$ preserve field-of-view.
Ground-truth correspondences are supplied at training time by the frozen
RoMa~v2 matcher~\cite{edstedt2026romav2}; pairs with fewer than $\tau_{\min}$ confident
matches are discarded.
The objective is
$\mathcal{L}_{\mathrm{corr}} = \|\hat{\mathbf{c}}_B - \mathbf{c}_B^*\|_1$.
Evaluation: $\mathrm{pct}@\{5,10\}\,\mathrm{px}$ in pixels and Average Pixel 
Error (APE).

\paragraph{(T2) Visual Localization.}
Given $\numviews$ reference views with known poses and a single unposed query
image, a visual localization head 
\texttt{VL}$(\scenetokens_{\mathrm{ref}},\,\scenetokens_{\mathrm{tgt}})$
regresses the absolute target pose as a 6D rotation~\cite{zhou2019continuity}
and 3D translation, all tokens extracted under identity poses.
Let $\mathcal{H}_\beta(a,b) = \mathrm{SmoothL1}(a,b,\beta)$; the loss is
\begin{equation}
  \mathcal{L}_{\mathrm{loc}}
  = \,\mathcal{H}_{0.05}(R_{\mathrm{pred}},\,R^*)
  + 0.5 \mathcal{H}_{0.05}(t_{\mathrm{pred}},\,t^*).
  \label{eq:loc-loss}
\end{equation}
Evaluation: AUC at $\{20^\circ,\, 30^\circ\}$, where joint AUC integrates $\min(\mathrm{recall}_R,\,\mathrm{recall}_t)$ normalized to the threshold range.

\paragraph{(T3) Multi-View Pose Estimation.}
Given $\numviews$ images in an \emph{unknown} coordinate frame, a pose estimation sequence head
\texttt{PES}$(\scenetokens)$ uses $\numviews$ learned
view queries to simultaneously regress all poses in a single forward pass.
Per-view validity indicators $m_k \in \{0,1\}$ mask padded entries:
\begin{equation}
  \mathcal{L}_{\mathrm{seq}}
  = \frac{1}{\sum_k m_k}
    \sum_{k=1}^{\numviews} m_k
    \bigl(
      \,\mathcal{H}_{0.05}(R_k,\,R_k^*)
      + 0.5 \mathcal{H}_{0.05}(t_k,\,t_k^*)
    \bigr).
\end{equation}
Metrics are identical to T2, marginalized over all valid views.

\paragraph{(T4) Pointmap and Depth Estimation.}
To test dense 3D scene structure, a DPT~\cite{ranftl2021dpt} readout head predicts relative depth for each patch token.
Evaluated on Hypersim~\cite{roberts2021mlhypersim} and 7Scenes~\cite{shotton2013scene} using $L_1$ relative depth error (AbsRel) and $\delta_1$, the percentage of pixels
satisfying $\max(\hat{d}/d, d/\hat{d}) < 1.25$.

\paragraph{(T5) Robot Manipulation.}
To evaluate viewpoint robustness in embodied settings, frozen encoder tokens serve as the visual state for an imitation learning policy. Evaluated on MimicGen~\cite{mandlekar2023mimicgen} under systematic camera shifts ($0.0$ to $1.0$ radians), measuring task success rate.

\section{Analysis: Decoder Expressivity and Prediction Target}

Our objective is to learn scene representations that transfer across geometric
tasks, not simply to optimize novel-view reconstruction. These goals can diverge:
an encoder-decoder NVS model~\cite{jin2025lvsm,
szymanowicz2026lagernvs} may reconstruct target views
accurately while placing little transferable spatial structure in the multi-view
encoder. We identify two key design factors that influence whether the learned
multi-view representation encodes transferable spatial structure: decoder
expressivity and the prediction target.

\subsection{Axis 1: Decoder Expressivity Degrades Geometric Transfer}
\label{sec:axis1}

\noindent \textbf{Decoder expressivity.}
In encoder-decoder NVS models, the decoder's capacity dictates the balance between its own learned 2D priors and the encoded conditioning signal. A highly expressive decoder acts as a strong unconditional prior, utilizing local spatial context to reconstruct or smooth images independently. While this improves raw synthesis quality, it creates an optimization shortcut: the strong prior weakens the model's reliance on the conditioning signal, relieving the encoder of the need to capture precise, multi-view geometry. Consequently, the downstream utility of the learned representation degrades. In transformer-based architectures, the strength of this prior is primarily controlled by the number of decoder layers and the receptive field of target-view self-attention.

Prior NVS models commonly use highly expressive decoders with full self-attention in the reconstruction of the target-view 
\cite{jin2025lvsm,szymanowicz2026lagernvs}.
We study the influence of decoder expressivity by varying the receptive field of these self-attention blocks.

\noindent \textbf{Experiment} To isolate this factor, we train \method\ variants with varying decoder
receptive fields, $\rf \in \{1{\times}1, 2{\times}2, 4{\times}4,
8{\times}8, 16{\times}16\}$. We control the receptive field using a
\blocknn\ mask, which restricts the decoder's self-attention to non-overlapping $k{\times}k$ spatial windows. Expanding the mask increases the amount of target-view spatial context the decoder can aggregate, while leaving the decoder architecture otherwise unchanged. At the strictest setting,
$\rf = 1{\times}1$, spatial communication between target queries is eliminated,
reducing the decoder to a patchwise residual MLP. At the broadest setting,
$\rf = 16{\times}16$, the window spans the full feature map, giving the decoder full self-attention in target-view, as in \cite{jin2025lvsm, szymanowicz2026lagernvs}. 

\begin{figure}[!t]
  \centering
  \includegraphics[width=\linewidth]{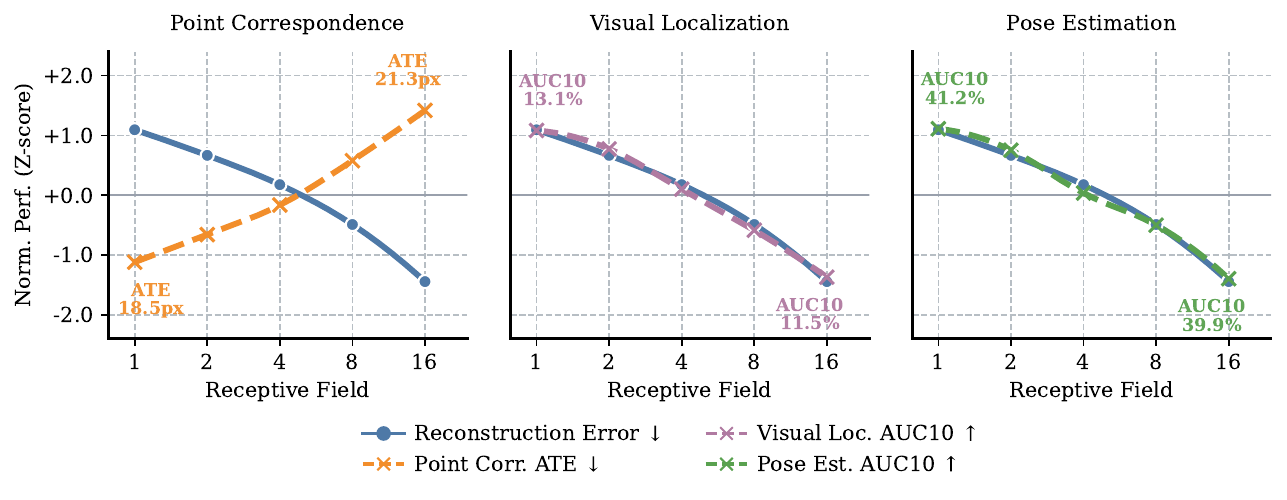}
  \vspace{-1em}
  \caption{
    \textbf{Effect of Receptive field across three geometric tasks on RE10K.}
    Normalized performance (Z-score) vs.\ $\rf$ on a $\log_2$ scale.
    Blue circles: reconstruction error (lower is better); dashed colored lines: downstream task performance (higher is better).
    \textbf{Left:} Point correspondence reveals the critical crossover: as $\rf$ expands, reconstruction error falls while correspondence error \textit{rises}.
    The decoder's improving reconstruction comes entirely at the cost of encoder spatial structure.
    \textbf{Center \& Right:} Visual localization and pose estimation performance degrade monotonically as $\rf$ increases, implying that as decoder expressivity increases, downstream representational performance decreases.
  }
  \label{fig:rf_ablation}
\end{figure}

\begin{figure}[!b]
    \centering
    \includegraphics[width=\linewidth]{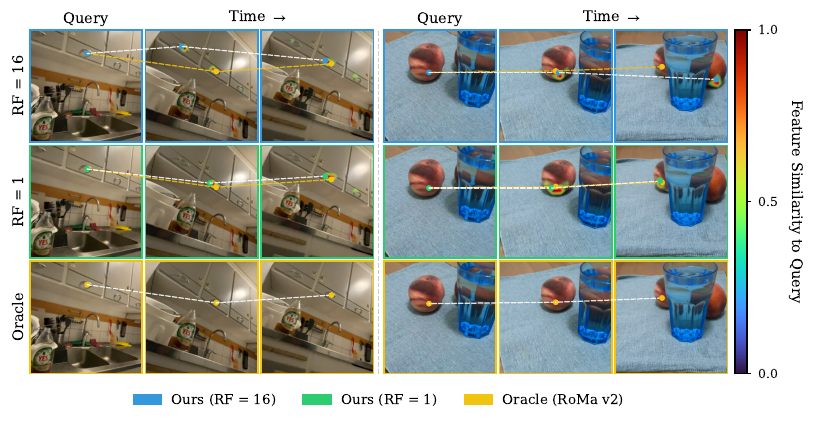}
    \caption{
    \textbf{Receptive field qualitative comparison.}
    Each row tracks a query point (colored dot) and its feature similarity map across time. While a large receptive field ($\rf = 16{\times}16$) yields more scattered similarity maps, restricting the receptive field to $\rf = 1{\times}1$ captures richer and more precise geometric correspondences albeit slightly more diffuse heatmaps, tracking with the Oracle across both indoor (kitchen, left) and object (peaches, right) scenes.
    }
    \label{fig:rf_qualitative}
\end{figure}

Figure~\ref{fig:rf_ablation} shows that increasing decoder receptive field improves the reconstruction objective but degrades downstream geometric transfer. 
As $\rf$ grows, the decoder is able to use more target-view spatial context, leading to lower feature reconstruction error. However, the frozen encoder representation becomes less useful for geometry: point correspondence error, visual localization, and pose estimation all degrade.
This inverse relationship suggests that reconstruction quality alone is a poor proxy for representation quality in encoder-decoder NVS models. The qualitative results in Figure~\ref{fig:rf_qualitative} show the same trend. With $\rf=1{\times}1$, similarity maps are sharply localized around the corresponding point and remain stable across viewpoints, while with complete self-attention ($\rf=16{\times}16$), the maps do not always track the ground truth correspondence. These results demonstrate limiting target-view spatial communication prevents the decoder from absorbing reconstruction priors, encouraging transferable spatial structure into the multi-view encoder, motivating our patch decoder.

\subsection{Axis 2: Prediction Space Determines Representation Quality}
\label{sec:axis2}

The prediction target determines what information the model is pressured to encode. 
Recent NVS models typically supervise reconstruction in pixel space~\cite{jin2025lvsm,
szymanowicz2026lagernvs}, which can emphasize low-level appearance, texture, and illumination rather than transferable geometry. 
We instead use latent feature prediction, where earlier ViT layers remain closer to RGB while deeper layers yield high-level structural abstractions that are robust to non-geometric artifacts like lighting and aliasing.

We sweep both the \dinovthree layer that is used as the featurized input and the layer used as the prediction target, asking which feature spaces are optimal for novel-view prediction, both at the input and at the target. We evaluate the resulting encoders on RE10K and DL3DV (val subset), reporting feature reconstruction error, nearest-neighbor point correspondence (PCK@0.1), patch retrieval (MRR),
and intrinsic dimensionality using the TwoNN estimator~\cite{Facco_2017}. 

\begin{figure}[!t]
  \centering
  \includegraphics[width=\linewidth]{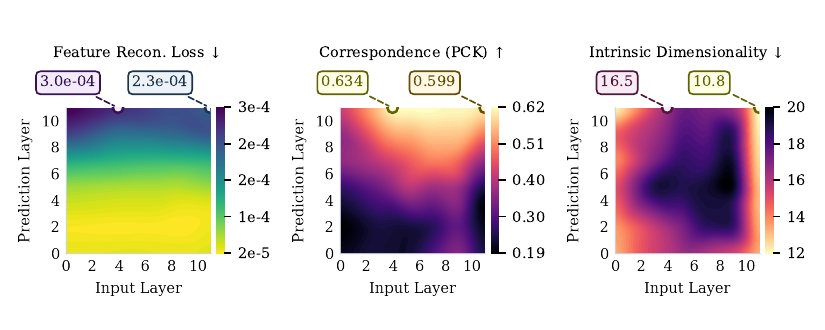}
  \vspace{-2em}
  \caption{
    \textbf{Performance landscape: initialization layer vs.\ prediction layer.}
    Each cell represents a model where the encoder outputs from the ``Input Layer'' and the decoder predicts features from the ``Prediction Layer'' of a 12-layer \dinovthree VIT-B (layers 0-11).
    Color encodes performance across the evaluated metrics, scaled such that yellow always indicates better performance.
    For downstream geometric tasks like correspondence (middle), predicting mid-to-high-layer targets generally yields the strongest relative performance.
    Conversely, predicting raw RGB or early-layer targets (bottom rows) consistently degrades these downstream geometric metrics.
    Notably, downstream representational capacity does not perfectly align with intrinsic dimensionality (right) or feature reconstruction (left) which is the validation cosine similarity; while RGB targets yield poor geometric features, they maintain a lower intrinsic dimensionality than several deep-layer prediction configurations.
  }
  \label{fig:landscape}
\end{figure}

Figure~\ref{fig:landscape} shows that the easiest reconstruction targets do not necessarily produce the best spatial representations. Lower-layer targets are closer to RGB and are easier to predict, with the best reconstruction error obtained by the (input: 9, target: 2) pairing. 
However, downstream geometry improves when predicting higher-layer DINOv3 features, suggesting that \emph{more semantic latent targets provide a better supervisory signal for transferable spatial structure}. 
The sweep also reveals a trade-off in the input feature space: input layer 4 gives the best in-distribution PCK and MRR, improving over layer 11 by roughly $5.8\%$ and $5.2\%$, respectively, but its representation has $\sim53\%$ higher intrinsic dimensionality.  We therefore use layer 11 as input, accepting a small loss in in-distribution accuracy for a more compressed and invariant multi-view representation. A comprehensive analysis of these trade-offs across all layers and metrics is provided in Appendix~\ref{sec:appendix_landscape}.

\section{Experiments}

We evaluate \method\ across five tasks using frozen encoder features with lightweight task-specific decoder heads. This protocol isolates the intrinsic spatial structure learned during pretraining, testing out-of-the-box generalization without confounding effects of task-specific fine-tuning data.

\noindent \textbf{Baselines.} We compare to different kinds of methods: \textbf{feed-forward NVS models} without any intermediate geometry prediction (LVSM \cite{jin2025lvsm}, LagerNVS \cite{szymanowicz2026lagernvs}, RayZer \cite{jiang2025rayzer}), \textbf{feed-forward 3DGS} prediction models (E-RayZer \cite{zhao2026erayzer}), and feed-forward \textbf{models trained for geometry prediction} itself (VGGT \cite{wang2025vggt}). We also compare to recent work on learning \textbf{multi-view representations using masked modeling} methods (Muskie  \cite{li2025muskiemultiviewmaskedimage}). 

To evaluate domain transfer, we split our protocol: point correspondence and relative pose are evaluated on the validation splits of our pretraining datasets (RealEstate10K~\cite{zhou2018stereo}, DL3DV~\cite{ling2024dl3dv}), while depth estimation and robot manipulation are evaluated on domains unseen by the vision backbone (Hypersim, 7Scenes, MimicGen) to evaluate out-of-domain generalization.

For perceiver-style NVS methods whose latents are not spatially organized (LVSM, RayZer), we cross-attend their latents to a shared $H{\times}W$ spatial grid and compute all downstream features from the resulting spatial grid tokens, ensuring a fair comparison with spatially indexed encoder methods.

Importantly, despite employing full cross-view attention, \method maintains a significantly smaller memory footprint than alternative models (Appendix \ref{sec:appendix_inference}).
By operating on $256{\times}256$ images with a patch size of 16, our encoder processes a sequence length of only $256\numviews + 8$ tokens (including registers).
In contrast, VGGT processes $518{\times}518$ images at a patch size of 14, resulting in $1369\numviews + 5$ tokens; it would require more than 12 views for our model to remotely approach the attention cost of these baselines.
Combined with Grouped Query Attention (GQA) configured with 16 heads and 8 groups, \method alleviates KV costs, operating efficiently even under limited compute budgets.

\subsection{Point Correspondence}

Point correspondence tests spatial discrimination at the pixel level: a match is correct only if the feature at one pixel is more similar to its true correspondent than to all other image pixels. Ground-truth correspondences for evaluation are supplied by a state-of-the-art dense matcher (RoMa~v2~\cite{edstedt2026romav2}) acting as a correspondence oracle. As Table~\ref{tab:correspondence} shows, \method\ outperforms self-supervised baselines on RealEstate10K across all pixel thresholds. 
Remarkably, \method\ achieves this higher sub-patch precision despite processing inputs at half the resolution of VGGT ($256{\times}256$ vs. $518{\times}518$), underscoring the spatial sharpness of our learned representations.
On DL3DV, where scene complexity is higher, \method\ performs competitively with VGGT despite utilizing strictly no dense 3D supervision, approaching its performance within fractional margins across all distance thresholds.
NVS baselines (ERayZer, LVSM, RayZer) consistently underperform, confirming that high reconstruction fidelity does not imply spatial feature discriminability.

\begin{table*}[t]
  \centering
  \small
  \setlength{\tabcolsep}{4.5pt}
  \caption{
    \textbf{Point correspondence.} 
    Pixel-level matching precision (PCK) and average trajectory/pixel error (APE). 
    Top results are highlighted in \textbf{bold}. 
    \method\ outperforms baselines on RE10K and remains competitive on DL3DV and Hypersim without explicit 3D supervision.
  }
  \label{tab:correspondence}
  \begin{tabular}{l ccc ccc ccc}
    \toprule
    & \multicolumn{3}{c}{RealEstate10K} & \multicolumn{3}{c}{DL3DV} & \multicolumn{3}{c}{Hypersim} \\
    \cmidrule(lr){2-4} \cmidrule(lr){5-7} \cmidrule(lr){8-10}
    & \multicolumn{2}{c}{PCK $\uparrow$} & APE $\downarrow$ 
    & \multicolumn{2}{c}{PCK $\uparrow$} & APE $\downarrow$ 
    & \multicolumn{2}{c}{PCK $\uparrow$} & APE $\downarrow$ \\
    \cmidrule(lr){2-3} \cmidrule(lr){5-6} \cmidrule(lr){8-9}
    Method & @5px & @10px & & @5px & @10px & & @5px & @10px & \\
    \midrule
    LVSM              &  5.51 & 14.76 & 41.04 & 3.53 & 11.05 & 45.75 & 5.62 & 13.41 & 42.15 \\
    \rowcolor[HTML]{EFEFEF}
    RayZer            &  3.81 & 13.53 & 39.63 & 2.73 & 13.53 & 47.03 & 4.21 & 11.21 & 44.82 \\
    ERayZer           &  4.59 & 12.47 & 39.67 & 3.55 & 11.26 & 45.30 & 4.31 & 14.89 & 40.54 \\
    \rowcolor[HTML]{EFEFEF}
    LagerNVS          &  9.61 & 27.36 & 26.54 & 3.81 & 11.31 & 42.41 & 8.94 & 24.30 & 31.02 \\
    Muskie            & 12.09 & 34.43 & 20.90 & 3.58 & 11.39 & 41.78 & 12.71 & 34.15 & 21.15 \\
    \rowcolor[HTML]{EFEFEF}
    VGGT              & 12.85 & 32.78 & 21.79 & 4.15 & \textbf{13.65} & 39.34 & 14.92 & \textbf{35.38} & \textbf{19.14} \\
    \midrule
    \textbf{\method}  & \textbf{15.48} & \textbf{36.29} & \textbf{18.41} & \textbf{4.20} & 13.33 & \textbf{39.26} & \textbf{15.01} & 34.87 & 19.82 \\
    \bottomrule
  \end{tabular}
\end{table*}

\subsection{Visual Localization}

Visual localization requires matching a query image against a reference map and estimating absolute camera pose, thereby combining per-patch discriminability with global geometric consistency under large viewpoint change.
We evaluate by training a lightweight regression head on frozen encoder features to directly output a 6D rotation matrix and 3D translation.
As Table~\ref{tab:pose_estimation} shows, \method\ establishes a new frontier among self-supervised models on most metrics, confirming that the structured features induced by receptive field restriction generalize beyond local correspondence to global pose inference.

\subsection{Relative Pose Estimation}

Relative pose estimation uses a direct regression head over the full encoder latent, including register tokens, with early-stage cross-view fusion.
On RealEstate10K (Table~\ref{tab:pose_estimation}), \method\ achieves high rotation accuracy among NVS-based methods and is competitive with other self-supervised and NVS methods on translation accuracy.
The result confirms that \method's encoder retains global geometric reasoning simultaneously with the local discriminative structure demonstrated in Table~\ref{tab:correspondence} --- the local decoder bottleneck does not degrade large-scale geometric inference.

\begin{table*}[!b]
  \centering
  \small
  \setlength{\tabcolsep}{3.0pt}
  \caption{
    \textbf{Visual localization and relative pose estimation} performance. 
    Metrics include Rotation Accuracy at $5^\circ$ ($R_5$), Translation Accuracy at $5^\circ$ ($T_5$), and AUC at $20^\circ$ ($A_{20}$) and $30^\circ$ ($A_{30}$). 
    Top results are highlighted in \textbf{bold}. 
    \method\ demonstrates superior geometric precision among self-supervised and NVS methods.
  }
  \label{tab:pose_estimation}

  \resizebox{\textwidth}{!}{%
  \begin{tabular}{l cccc cccc | cccc cccc}
    \toprule
    & \multicolumn{8}{c}{Visual Localization (AUC / Acc $\uparrow$)}
    & \multicolumn{8}{c}{Relative Pose Estimation (AUC / Acc $\uparrow$)} \\
    \cmidrule(lr){2-9} \cmidrule(lr){10-17}

    & \multicolumn{4}{c}{RE10K}
    & \multicolumn{4}{c}{DL3DV}
    & \multicolumn{4}{c}{RE10K}
    & \multicolumn{4}{c}{DL3DV} \\
    \cmidrule(lr){2-5} \cmidrule(lr){6-9} \cmidrule(lr){10-13} \cmidrule(lr){14-17}

    Method
    & R$_5$ & T$_5$ & A$_{20}$ & A$_{30}$
    & R$_5$ & T$_5$ & A$_{20}$ & A$_{30}$
    & R$_5$ & T$_5$ & A$_{20}$ & A$_{30}$
    & R$_5$ & T$_5$ & A$_{20}$ & A$_{30}$ \\
    \midrule

    LVSM
    & 69.87 & 8.48 & 20.62 & 29.39
    & 20.39 & 3.55 & 6.02 & 11.82
    & 70.31 & 38.98 & 43.03 & 46.97
    & 54.41 & 32.37 & 27.65 & 30.18 \\
    \rowcolor[HTML]{EFEFEF}
    RayZer
    & 56.30 & 7.29 & 13.74 & 20.10
    & 22.85 & 2.92 & 5.13 & 10.01
    & 78.32 & \textbf{42.89} & 53.00 & 59.60
    & 58.32 & 34.91 & 32.11 & 36.10 \\

    ERayZer
    & 59.09 & 6.65 & 15.47 & 23.53
    & 24.04 & 3.05 & 5.99 & 12.03
    & 77.82 & 40.45 & 46.63 & 54.07
    & 50.82 & 31.29 & 23.56 & 27.32 \\
    \rowcolor[HTML]{EFEFEF}
    LagerNVS
    & 58.65 & 9.42 & 17.07 & 25.01
    & 22.81 & 7.35 & 5.79 & 12.05
    & 70.81 & 30.74 & 35.67 & 42.17
    & 57.81 & 36.91 & 34.96 & 44.13 \\

    Muskie
    & 79.65 & 11.50 & 23.25 & 32.56
    & 23.02 & 8.54 & 6.90 & 13.85
    & 80.26 & 40.09 & 44.63 & 53.07
    & 65.36 & 38.38 & 47.54 & 54.64 \\
    \rowcolor[HTML]{EFEFEF}
    VGGT
    & 74.31 & 9.58 & 25.01 & 34.70
    & 22.98 & \textbf{11.83} & \textbf{8.86} & \textbf{17.36}
    & \textbf{82.15} & 42.36 & \textbf{59.36} & \textbf{68.74}
    & \textbf{74.25} & \textbf{44.48} & \textbf{59.32} & \textbf{67.94} \\

    \midrule

    \textbf{SNAP}
    & \textbf{80.32} & \textbf{15.47} & \textbf{26.41} & \textbf{34.72}
    & \textbf{24.62} & 8.59 & 7.58 & 15.91
    & 78.81 & 41.42 & 53.07 & 62.30
    & 65.31 & 39.91 & 49.31 & 57.43 \\

    \bottomrule
  \end{tabular}%
  }
\end{table*}

\subsection{Robot Manipulation Under Viewpoint Shift}
\begin{figure}
  \centering
  \includegraphics[width=1.0\linewidth]{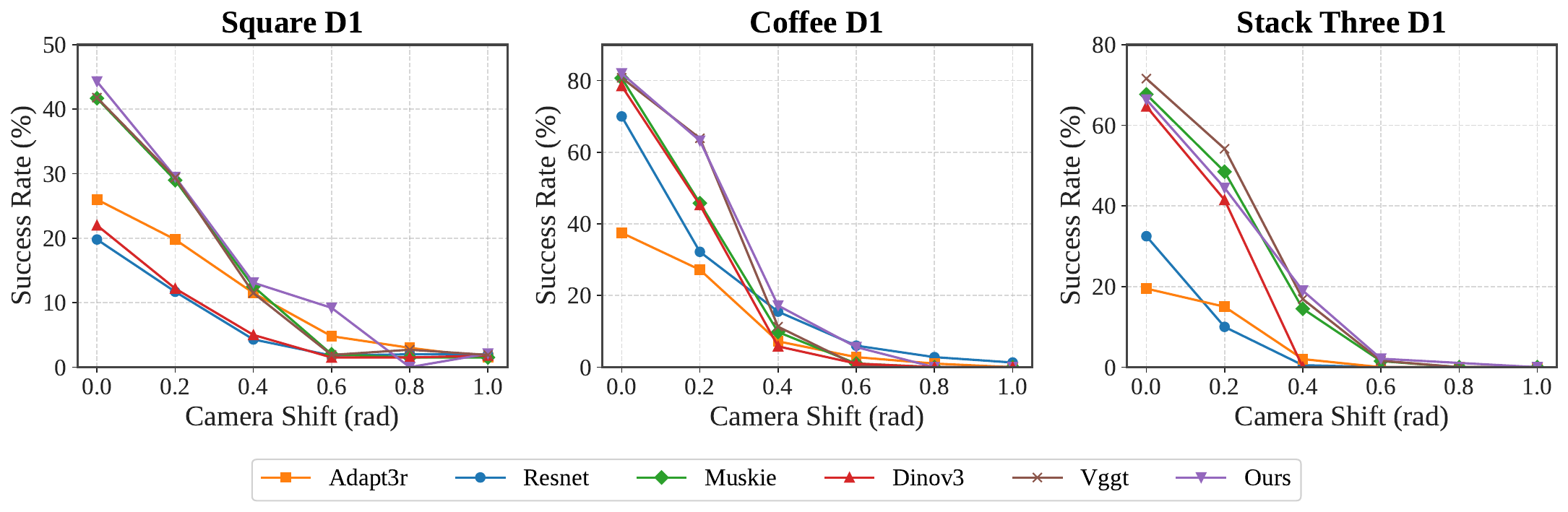}
  \vspace{-15pt}
  \caption{
      \textbf{Emergent Viewpoint Invariance in Robot Manipulation.} Success rate across three MimicGen tasks evaluated under camera shifts ($0.0$ to $1.0$ radians). While standard 2D semantic priors (DINOv3, ResNet~\cite{he2016resnet}) collapse rapidly under viewpoint perturbation, \method\ (Ours) maintains highly robust spatial representations, tracking closely with the geometry-supervised VGGT ceiling and systematically outperforming methods explicitly designed for spatial robustness (Adapt3R).
  }
  \label{fig:MimicGen}
\end{figure}

Beyond static benchmarks, we evaluate whether \method can transfer effectively to online decision-making. We compare \method against a representative suite of baselines, including semantic backbones (DINOv3, ResNet), geometry-supervised encoders (VGGT), and special-purpose pipelines (Adapt3R \cite{wilcox2025adapt3r}) on robotic manipulation under out-of-distribution viewpoint shifts. We evaluate on three MimicGen tasks under camera shifts of $0.0$ to $1.0$ radians using frozen encoder backbones. Despite not being trained on any robotics-related data, \method is competitive across the board.

As Figure~\ref{fig:MimicGen} shows, semantic encoders (DINOv3, ResNet) 
collapse rapidly beyond $0.2$ radians. Arguably, 2D contrastive training 
suppresses the cross-view spatial variations which the policy requires to infer 3D configuration. 
Conversely, \method's multi-view latent reconstruction 
objective induces strong viewpoint invariance out-of-the-box. Notably, \method\ outperforms its own frozen \dinovthree\ teacher, generally outperforms task-specific baseline Adapt3R, and tracks closely with the fully geometry-supervised VGGT baseline across all tasks and shift magnitudes.

\subsection{Relative Depth Estimation}

Relative depth estimation requires predicting scale-invariant depth for each image patch, a direct test of whether an encoder has internalized 3D scene structure. We train a depth head on Hypersim and evaluate performance on 7Scenes and Hypersim validation split. Notably, while LagerNVS was pre-trained on Hypersim and operates at a 518 resolution, LVSM and \method\ are evaluated zero-shot. Furthermore, LVSM utilizes a smaller $p=8$ patch size, providing a higher-resolution latent grid than \method.
As Table~\ref{tab:relative_depth} shows, \method\ outperforms NVS-based encoders. Using explicit 3D supervision (VGGT) performs better in this task. While exposure to Hypersim during pre-training yields high synthetic $\delta_1$ for LagerNVS, \method\ demonstrates strong generalization on the real-world 7-Scenes dataset. We supply qualitative results in Appendix~\ref{sec:appendix_qualitative_depth}.

\begin{table}[!h]
  \centering
  \begin{minipage}[c]{0.38\textwidth}
    \caption{
      \textbf{Relative depth estimation.}
      We evaluate relative depth prediction on a validation-split of Hypersim, and 7Scenes, a real-world held-out dataset.
      We report AbsRel and $\delta_1$, the percentage of pixels satisfying
      $\max(\hat{d}/d, d/\hat{d}) < 1.25$.
      Reporting for VGGT comes from the off-the-shelf depth estimation head.
    }
    \label{tab:relative_depth}
  \end{minipage}
  \hfill
  \begin{minipage}[c]{0.60\textwidth}
    \centering
    \small
    \begin{tabular}{l cc cc}
      \toprule
      & \multicolumn{2}{c}{Hypersim} & \multicolumn{2}{c}{7Scenes} \\
      \cmidrule(lr){2-3} \cmidrule(lr){4-5}
      Method & AbsRel $\downarrow$ & $\delta_1$ $\uparrow$
             & AbsRel $\downarrow$ & $\delta_1$ $\uparrow$ \\
      \midrule
      \textit{VGGT}      & \textit{0.2080} & \textit{94.6036}
                         & \textit{1.9138} & \textit{70.0987} \\
      \midrule
      LVSM               & 2.0648          & 38.5086
                         & 1.9474          & 32.5691 \\
      LagerNVS           & 0.7441          & \textbf{73.1069}
                         & 2.3858          & 49.2679 \\
      ERayZer            & 1.1321          & 50.6075 
                         & 1.3320          & 35.5672 \\
      \midrule
      \rowcolor[HTML]{EFEFEF}
      \textbf{\method\ (ours)}
                         & \textbf{0.7093} & 64.6815
                         & \textbf{0.9257} & \textbf{52.0218} \\
      \bottomrule
    \end{tabular}
  \end{minipage}
\end{table}

\section{Conclusion}
While our current pretraining relies on posed video sequences and static, textured scenes, and our findings are bound by a fixed compute and data budget that has yet to reveal the full scaling laws of the architecture, SNAP demonstrates that feed-forward novel view synthesis (NVS) is a legitimate powerhouse for self-supervised geometric pretraining. Extending this paradigm to dynamic video remains a critical next step in establishing this approach as a foundation model. More broadly, our results show that NVS's rich geometric signal is most available when two architectural pathologies are corrected: decoders expressive enough to absorb spatial reasoning that should belong to the encoder, and pixel-space targets that contaminate gradients with low-level statistics.

Restricting decoder receptive field via a \blocknn mask biases spatial reasoning back into the scene tokens; replacing pixel targets with last-layer DINOv3 features further redirects optimization toward a structured, geometry-privileged manifold. The resulting backbone rivals geometry-supervised models across five tasks without dense 3D annotations, and transfers robustly to robot manipulation, where viewpoint shifts break standard 2D semantic models. Moreover Appendix~\ref{sec:appendix_teachers} suggests that these improvements are not restricted to a single teacher, improving correspondence among \dinovthree, SAM, and MAE alike. Most strikingly, \method demonstrates NVS's geometric signal is best leveraged by restricting decoder capacity, not by optimizing reconstruction, underscoring that representation quality does not always correspond with reconstruction fidelity.
\newpage

{
\small
\bibliography{references}
}
\newpage

\appendix
\section*{Supplementary Material}
\addcontentsline{toc}{section}{Appendix}

\section{Related Work}
\label{sec:appendix_related_work}
\subsection{Feed-forward novel-view synthesis}

Novel-view synthesis has traditionally been addressed through per-scene
optimization of explicit or implicit 3D representations, including neural
radiance fields and 3D Gaussian splatting
\cite{mildenhall2020nerf,mueller2022instant,kerbl3Dgaussians}. These methods
produce high-quality renderings, but require optimizing a new scene
representation at test time. A parallel line of work instead relies on reconstruction priors learned from data, using feed-forward models for generalizable view
synthesis. Early examples condition radiance fields or image-based rendering
networks on sparse input views
\cite{yu2021pixelnerf,wang2021ibrnet,chen2021mvsnerf}, while recent methods
directly predict feed-forward 3D Gaussian scene representations
\cite{charatan2024pixelsplat,chen2024mvsplat,szymanowicz2024splatter}. More recently, large transformer-based NVS models have moved further away from
3D representations, synthesizing target views directly from learned
multi-view representations or scene tokens
\cite{jin2025lvsm,jiang2025rayzer,
zhao2026erayzer,mitchel2026xfactor,
szymanowicz2026lagernvs,kim2026svsm}. These models are
typically evaluated by image reconstruction quality, but their learned
intermediate representations have also been proposed as general-purpose
multi-view scene representations. Our work investigates aspects of NVS pretraining to improve the learned representation for downstream tasks, rather than only for reconstructing novel views.

\subsection{Multi-view representation learning}

Beyond novel-view synthesis, several works learn multi-view representations
explicitly for geometric reasoning. CroCo introduced cross-view completion as a self-supervised pretraining objective for two-view 3D vision \cite{weinzaepfel2022croco}. This representation was later used as the foundation for DUSt3R and MASt3R, which used pointmap prediction to solve many diverse tasks  \cite{wang2024dust3r,leroy2024mast3r}. Recent multi-view generalist models such as VGGT, CUT3R, $\pi^3$, MapAnything, D4RT, and Flow3R further extend this direction, predicting camera parameters, depth, pointmaps, tracks, or dynamic scene structure from arbitrary image collections \cite{wang2025vggt,wang2025cut3r, wang2026pi3,keetha2026mapanything, zhang2026d4rt,cong2026flow3r}. Many of the strongest geometric foundation models rely on explicit 3D supervision, such as depth, pointmaps, camera poses, or tracks, which limits the scale and diversity of usable training data. Self-supervised alternatives such as Muskie and MuM \cite{li2025muskiemultiviewmaskedimage,nordstrom2025mum} revisit masked modeling in the multi-view setting. Our work is complementary: rather than using masked reconstruction as the pretext task, we study novel-view prediction as a scalable self-supervised signal, and identify how decoder expressivity and prediction target determine whether NVS produces transferable spatial representations.

\subsection{3D awareness in video and multi-view models}

Recent work studies whether pretrained 2D, video, and multi-view foundation
models encode useful 3D structure. \cite{elbanani2024probe3d}
probes frozen 2D visual foundation models for single-view geometry and cross-view
surface consistency, finding that some self-supervised image features encode
non-trivial 3D information. Feat2GS~\cite{chen2025feat2gs} further probes visual
foundation models by reading out 3D Gaussian attributes from frozen features,
using novel-view synthesis to evaluate geometry and texture awareness. VidFM3D~\cite{huang20253dvideofoundationmodels} shows that video foundation model
features also contain 3D-aware information extractable with lightweight geometry
heads. Our work asks a complementary question: what makes NVS pretraining produce transferable spatial representations? Rather than only
probing for 3D information after pretraining, we isolate two design factors
that shape the learned encoder representation: decoder expressivity and the
prediction target. We also evaluate our representation on a range of geometric and robotics tasks. 

Among these, the architectures closest to ours are SVSM~\cite{kim2026svsm},
LagerNVS~\cite{szymanowicz2026lagernvs}, and
D4RT~\cite{zhang2026d4rt}, all concurrent works with distinct objectives.
SVSM and LagerNVS both follow prior work in pairing the scene representation
with an expressive decoder, which is the main architectural difference from
our approach; as we show in Sec. ~\ref{sec:axis1}, this design choice likely limits how
well their learned representations transfer to downstream multi-view tasks.
Within this pair, SVSM focuses on the data scale required for pretraining,
while LagerNVS instead focuses on distilling priors from geometry-pretrained
models such as VGGT. D4RT, in contrast, adopts a lightweight cross-attention
decoder similar to ours, but trains it with supervised geometry targets,
whereas we rely on a purely self-supervised NVS objective.

\section{Implementation Details}
\label{sec:appendix_methods}

\subsection{Pretraining Setup}

\textbf{Datasets and Splits.} We train \method\ on a large-scale mixture of multiview video datasets totaling approximately 117,000 sequences. This training set consists of the RealEstate10K training partition ($\sim$67,000 sequences), DL3DV-10K ($\sim$10,000 sequences), and Co3Dv2 ($\sim$40,000 sequences). For the DL3DV and Co3Dv2 datasets, we hold out 20\% of the available sequences for validation, and we use the preprovided test split for validation on RE10K. During training, we randomly sample up to $N=6$ reference views and one target view per sequence. All images are center-cropped and resized to $256 \times 256$ pixels. We use 16 layers of self attention for the multi-view encoder with embedding dimension of 1024 and 8 blocks of alternating \prope \cite{li2025cameras} cross-attention and masked self attention. We leverage grouped-query attention \cite{ainslie2023gqa} operating with 16 query heads and 8 KV-groups.

\textbf{Optimization.} Training was conducted for 40,000 steps using an effective batch size of 256 across multiple classes of GPU hardware using Distributed Data Parallel (DDP). We employ the AdamW optimizer with a peak learning rate of $1 \times 10^{-4}$ and a weight decay of 0.02. The learning rate follows a linear warmup for the first 2,000 steps, followed by a cosine annealing schedule that decays toward zero. We apply gradient clipping with a maximum norm of 0.5 to maintain training stability.

\textbf{Exponential Moving Average (EMA).} To stabilize the learned representations, we maintain an Exponential Moving Average of all trainable parameters with a decay rate of 0.999. The EMA scope encompasses every component of the architecture, including the pose-free multi-view encoder and the pose-conditioned local decoder, which serves as the latent feature reconstruction predictor. All empirical evaluations and downstream geometric probes reported in the main text utilize these EMA weights.

\subsection{Point Correspondence}

\textbf{Label Generation.} Training labels are generated using image pairs from the train splits of RealEstate 10K, DL3DV-10K, and Hypersim. We employ two distinct strategies depending on the dataset:
\begin{itemize}
    \item \textbf{RE10K and DL3DV:} We derive pseudo-ground-truth correspondences from a frozen RoMa v2 matcher. To ensure high-quality supervision, we filter for image pairs containing at least $\tau_{min}=8$ confident matches.
    \item \textbf{Hypersim:} We compute exact ground-truth correspondences via cross-projection using provided depth maps and camera poses.
\end{itemize}

\textbf{Architecture.} The point correspondence head employs a three-stage mechanism: (1) symmetric cross-attention layers that map the encoder's latent tokens to a canonical $16\times16$ spatial grid; (2) a source-view cross-attention block to localize the discriminative feature at query coordinate $c_{A}$; and (3) a matching stage that computes angular similarity between the query and target-view features via $L_{2}$ normalized dot products. 

Specifically, the decode stage (1) consists of 3 cross-attention layers, and the localization stage (2) consists of 3 cross-attention layers. In the final matching stage (3), independent linear projections are applied to the queries and keys to prevent gradient interference. The $L_{2}$ normalized dot product is scaled by a learnable, log-parameterized temperature scalar (initialized to 1.0) before calculating the similarity heatmap. A differentiable spatial expectation over the resulting similarity heatmap produces the final predicted coordinate.

\textbf{Optimization.} The correspondence head is trained for 20,000 steps with an effective batch size of 128. We utilize the AdamW optimizer with a peak learning rate of $2\times10^{-4}$, a 2000-step linear warmup, and a cosine annealing schedule. As specified in our evaluation protocol, all views are encoded under identity camera-to-world poses with pixel-unit intrinsics to isolate the encoder's intrinsic geometric awareness from external pose signals. Evaluation occurs on the held-out validation splits on all datasets.

\subsection{Visual Localization}

\textbf{Architecture.} The visual localization head is designed to regress an absolute target pose by localizing an unposed query image against a reference multi-view context. The target pose is parameterized using 9 parameters: a continuous 6D rotation representation and a 3D translation. 

The regression head utilizes alternating transformer layers of self-attention (processing the target sequence) and cross-attention (reading from the reference context). A learned CLS token is prepended to the target image sequence, and 1D RoPE is applied to encode sequence positions. After processing the tokens through the transformer blocks, the refined CLS token is extracted and passed through a final multi-layer perceptron to output the 9 pose parameters. To ensure stable early training, the final linear layer's biases are initialized to represent an identity transformation.

\textbf{Optimization.} We train this head using the RealEstate 10K and DL3DV train splits, and evaluation is conducted on their respective held-out validation splits. Optimization uses a peak learning rate of $2\times10^{-5}$ with a cosine annealing schedule and 2,500 steps of linear warmup. The network minimizes a 9D pose regression loss using a SmoothL1 penalty on both the predicted rotation $R_{pred}$ and translation $t_{pred}$.

\subsection{Pose Estimation}

\textbf{Architecture.} The relative pose estimation head predicts the pose sequence for all $N$ input views simultaneously in a single forward pass. Similar to the visual localization head, this module uses transformer layers alternating between self-attention and cross-attention. It queries the multi-view context using $N$ learned view queries (one corresponding to each view) equipped with 1D RoPE. The output tokens are processed by a regression MLP (Linear $\rightarrow$ GELU $\rightarrow$ Linear) to predict the 9D pose (6D rotation and 3D translation) for each respective view. Per-view validity indicators mask out padded entries during the loss computation.

\textbf{Optimization.} The sequence head is trained on the RealEstate 10K and DL3DV train splits and evaluated on their held-out validation splits. Training utilizes a peak learning rate of $2\times10^{-5}$, governed by a cosine annealing schedule with 2,500 warmup steps. The model is optimized using a 9D pose regression loss based on the Smooth-L1 error of the rotation and translation predictions.

\section{Robot Manipulation}
\begin{figure}[t]
  \centering
  \includegraphics[
    width=1.0\linewidth,
    trim=0cm 7.5cm 8.5cm 9.5cm, 
    clip
  ]{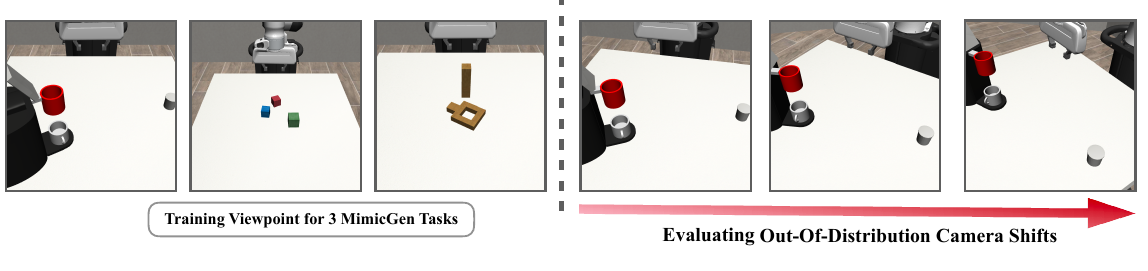}
  \vspace{-10pt}
  \caption{\textbf{Training and evaluation setting for robotics tasks}: \textbf{Left}: Three MimicGen simulation task we train our downstream policies on. \textbf{Right}: To test the model's emergent viewpoint invariance, we evaluate its performance on out-of-distribution camera shifts ranging from 0.2 to 1.0 radians, none of which were encountered during training.
  }
  \label{fig:MimicGen_overview}
\end{figure}

Here we discuss details about the sim experiments, including detailed information about the inputs to the policies, hyperparameter choices, etc.

\paragraph{Tasks.}
We evaluate on three MimicGen manipulation tasks: \textsc{Stack Three-D1}, \textsc{Square-D1}, and \textsc{Coffee-D1}. 
Each task is solved by a single 7-DoF Franka Panda arm with a parallel-jaw gripper. 
For every task we train on the full MimicGen demonstration set of $1{,}000$ demonstrations generated from the source human demos. 
Observations are RGB images from two cameras: a fixed third-person camera and a wrist-mounted camera on the robot. The policy is only conditioned on these visual features.

\paragraph{Vision Encoders.}
We benchmark six vision backbones with otherwise identical training and
policy hyperparameters. 
The encoders we benchmark are the following: (i) \textbf{ImageNet-pretrained ResNet-18}; (ii) \textbf{Adapt3R}, a 3D-aware encoder built on a CLIP RN50 backbone with a point-cloud attention extractor and pyramid feature network; (iii) \textbf{DINOv3} ViT-B/16; (iv)
\textbf{Muskie}, a multi-view backbone with
alternating attention; (v) \textbf{VGGT}, a geometry-aware transformer aggregator; (vi) \textbf{SNAP} (ours), a multi-view backbone built on a frozen DINOv3 featurizer.
For VGGT and SNAP we are using their patch, register, and camera (only applicable to VGGT) tokens. 
Note that all multi-view encoders see both cameras as input views ($V{=}2$), whereas single-view encoders (ResNet, Adapt3R, DINOv3) process each camera independently and their tokens are concatenated downstream. As for the policy, most encoders (Muskie, DINOv3, ResNet, SNAP) we use their last layer features to condition the policy. 
For VGGT, we use the element-wise mean of all four intermediate aggregator-layer features, preserving the per-token shape, which has been shown to an extent to provide better viewpoint robustness \cite{abouzeid2025geoaware}.

\paragraph{Policy.}
For all encoders we use the same diffusion policy, a Diffusion-Transformer (DiT) that predicts a chunk of $H_a{=}8$ future actions in delta pose representation, of which the first $H_e{=}4$ are executed (action-horizon) before re-planning. 
The policy uses $1$ encoder block and $3$ transformer decoder blocks with hidden dimension
$d_\text{embed}{=}512$, $4$ attention heads, GELU activations, and dropout $0.1$. 
The diffusion process uses $100$ training steps and $8$ DDIM~\cite{song2021ddim} sampling steps at evaluation. 
We train all models using AdamW ($\beta_1{=}0.95$, $\beta_2{=}0.999$) with a learning rate of $1\times 10^{-4}$ and weight decay of $1\times 10^{-6}$, a cosine schedule with $1{,}000$ warm-up steps decaying to $1\%$ of the peak, batch size $128$, and gradient clipping at $100$ for $350$ epochs. 
Models use embedding dimension $d_e{=}512$, chunk size $8$, and action horizon $4$. Results are reported over 400 rollouts per task per policy.

\paragraph{Image Resolution and Augmentation.}
We use a common training resolution of $256{\times}256$ for DINOv3,
SNAP.
For VGGT and Muskie we used a resolution of $224{\times}224$.
ResNet and Adapt3R are trained at $128{\times}128$ resolution (following a similar spirit as Adapt3r's training resolution). 
The same lightweight augmentation is applied to every encoder: per-batch color jitter (brightness, contrast, saturation, hue) followed by a random translation crop of $\pm 8$ pixels.

\section{Comparison with Teachers}
\label{sec:appendix_teachers}

\subsection{Comparison with Frozen Teachers}
\label{sec:appendix_frozen_teachers}

SNAP uses a frozen visual teacher to provide feature targets for its feature-space reconstruction objective. While our main experiments use DINOv3~\cite{simeoni2025dinov3}, the proposed multi-view pretraining objective is not specific to this teacher. We therefore compare SNAP using frozen DINOv3~\cite{simeoni2025dinov3}, SAM~\cite{Kirillov_2023_ICCV}, and MAE~\cite{he2022mae} features.

\paragraph{Point Correspondence.}
As shown in Table~\ref{tab:teacher_comparison}, SNAP substantially improves point correspondence for every teacher, dataset and threshold. 
\footnote{All point-correspondence numbers in this appendix are computed via training free methods and use global soft-argmax matching with a low temperature ($\tau=0.02$): each source keypoint is compared against the full target feature map, and the prediction is the soft-argmax of the resulting similarity map.
Results are aggregated over multiple random seeds, for roughly 2 million evaluated pairs in total. SNAP, \dinovthree, SAM and MAE use ViT-B backbones; V-JEPA 2.1 uses a ViT-L variant, a larger model that favors the baseline.}

On held-out Hypersim, PCK@0.1 improves from 29.29 to 55.35 with DINOv3~\cite{simeoni2025dinov3}, from 40.82 to 52.27 with SAM~\cite{Kirillov_2023_ICCV}, and from 11.01 to 50.67 with MAE~\cite{he2022mae}. 
The gains are largest at tight thresholds and shrink at PCK@0.3 for the stronger teachers (\dinovthree: 84.22 to 88.35; SAM: 78.74 to 83.64), which approach saturation, while MAE still improves by 23.35 points (60.81 to 84.16). 
The same trends hold on RealEstate10K and DL3DV.
Thus, the geometric gains are not specific to the \dinovthree feature space. Instead, the multi-view, pose-conditioned reconstruction objective consistently improves the spatial consistency and geometric utility of the underlying teacher representation.

\begin{table}[!t]
  \centering
  \small
  \setlength{\tabcolsep}{4.5pt}
  \caption{
    \textbf{Point correspondence with frozen teacher features.}
    PCK (\%) at thresholds $0.1$, $0.2$ and $0.3$ for frozen teacher features
    (Base) and the corresponding SNAP representations, using global
    soft-argmax matching ($\tau=0.02$). Hypersim is held out from SNAP's
    training. SNAP improves over its teacher for every teacher, dataset and
    threshold.
  }
  \label{tab:teacher_comparison}

  \begin{tabular}{llccccccccc}
    \toprule
    & & \multicolumn{3}{c}{RealEstate10K}
      & \multicolumn{3}{c}{DL3DV}
      & \multicolumn{3}{c}{Hypersim} \\
    \cmidrule(lr){3-5} \cmidrule(lr){6-8} \cmidrule(lr){9-11}
    Teacher & Features
    & @0.1 & @0.2 & @0.3
    & @0.1 & @0.2 & @0.3
    & @0.1 & @0.2 & @0.3 \\
    \midrule

    \multirow{2}{*}{\dinovthree~\cite{simeoni2025dinov3}}
    & Base & 45.83 & 80.13 & 93.71 & 28.00 & 62.53 & 84.45 & 29.29 & 63.67 & 84.22 \\
    & SNAP & \textbf{78.21} & \textbf{91.32} & \textbf{95.62} & \textbf{58.74} & \textbf{80.27} & \textbf{89.61} & \textbf{55.35} & \textbf{77.90} & \textbf{88.35} \\
    \midrule

    \rowcolor[HTML]{EFEFEF}
    & Base & 64.60 & 81.61 & 89.99 & 49.17 & 70.75 & 82.71 & 40.82 & 64.80 & 78.74 \\
    \rowcolor[HTML]{EFEFEF}
    \multirow{-2}{*}{SAM~\cite{Kirillov_2023_ICCV}}
    & SNAP & \textbf{75.58} & \textbf{87.40} & \textbf{92.81} & \textbf{60.09} & \textbf{77.53} & \textbf{86.58} & \textbf{52.27} & \textbf{72.62} & \textbf{83.64} \\
    \midrule

    \multirow{2}{*}{MAE~\cite{he2022mae}}
    & Base & 17.71 & 50.33 & 76.96 & 13.50 & 42.67 & 70.15 & 11.01 & 35.31 & 60.81 \\
    & SNAP & \textbf{75.34} & \textbf{89.02} & \textbf{94.17} & \textbf{59.64} & \textbf{78.19} & \textbf{87.21} & \textbf{50.67} & \textbf{72.81} & \textbf{84.16} \\

    \bottomrule
  \end{tabular}
\end{table}

Among the base teachers, \dinovthree is strongest at PCK@0.3, whereas SAM is stronger (or tied) at PCK@0.1 and PCK@0.2 on all three datasets (e.g., Hypersim PCK@0.1: 40.82 vs.\ 29.29). After SNAP, the spread between teachers narrows considerably (Hypersim PCK@0.1 ranges from 50.67 to 55.35, versus 11.01 to 40.82 for the base features), and \dinovthree yields the strongest SNAP representation in all but one setting (DL3DV PCK@0.1). \dinovthree is therefore a good choice but not a requirement for SNAP. The substantial improvements obtained with SAM~\cite{Kirillov_2023_ICCV} and MAE~\cite{he2022mae} demonstrate that the gains arise from combining high-order abstract representations with the view synthesis objective rather than simply inheriting the capabilities of the \dinovthree teacher. The remaining differences between teachers show that the quality of the underlying feature space is still significant.

\paragraph{Relative Depth Estimation.}
We further compare \method against its \dinovthree~\cite{simeoni2025dinov3} teacher in Table~\ref{tab:relative_depth_teacher_comparison}. \method outperforms its teacher on Hypersim, maintaining more accurate relative depth along with a better $\delta_1$. This pattern holds in 7Scenes where SNAP displays a 15.8\% improvement in accuracy in depth estimation while displaying a comparable $\delta_1$. The improvements demonstrate that SNAP exceeds its teacher at dense prediction tasks as well as point correspondence.

\begin{table}[!t]
  \centering
  \small
  \setlength{\tabcolsep}{5.5pt}
  \caption{
    \textbf{Relative depth estimation on Hypersim and 7Scenes.}
    AbsRel ($\downarrow$) and $\delta$ ($\uparrow$) for
    DINOv3~\cite{simeoni2025dinov3} and SNAP.
    Best results are highlighted in \textbf{bold}.
  }
  \label{tab:relative_depth_teacher_comparison}

  \begin{tabular}{lcccc}
    \toprule
    & \multicolumn{2}{c}{Hypersim}
    & \multicolumn{2}{c}{7Scenes} \\
    \cmidrule(lr){2-3} \cmidrule(lr){4-5}
    Method
    & AbsRel $\downarrow$ & $\delta_1$ $\uparrow$
    & AbsRel $\downarrow$ & $\delta_1$ $\uparrow$ \\
    \midrule

    DINOv3~\cite{simeoni2025dinov3}
    & 0.7147 & 61.8738
    & 1.0991 & \textbf{52.8681} \\

    \rowcolor[HTML]{EFEFEF}
    SNAP
    & \textbf{0.7093} & \textbf{64.6815}
    & \textbf{0.9257} & 52.0218 \\

    \bottomrule

  \end{tabular}
\end{table}

\paragraph{Single-View Semantics.}
We additionally evaluate whether the multi-view transformation preserves
single-view semantic information. As shown in
Table~\ref{tab:teacher_imagenet}, the DINOv3~\cite{simeoni2025dinov3}
variant improves slightly on ImageNet~\cite{deng2009imagenet} $k$-NN, from 68.45 to 68.81 Top-1
accuracy. The SAM~\cite{Kirillov_2023_ICCV} variant also improves, from
18.06 to 21.46 Top-1 accuracy, while the MAE~\cite{he2022mae} variant
decreases from 30.78 to 28.64. On ADE20K~\cite{zhou2017ade20k} linear-probe segmentation,
DINOv3~\cite{simeoni2025dinov3} obtains 51.92 mIoU and 87.27 pixel accuracy,
compared with 49.87 mIoU and 84.31 pixel accuracy for SNAP
(Table~\ref{tab:teacher_ade20k}). These results show that SNAP largely
retains the semantic discriminability of the teacher while deliberately
shifting the representation toward multi-view geometric transfer.

\begin{table}[!tbp]
  \vspace{-15pt}
  \centering
  \small
  \setlength{\tabcolsep}{5.5pt}
  \caption{
    \textbf{ImageNet $k$-NN classification with frozen representations.}
    Top-1 and Top-5 accuracy (\%) are reported for the original teacher
    representations and their corresponding SNAP representations.
    Best results are highlighted in \textbf{bold}.
  }
  \label{tab:teacher_imagenet}

  \begin{tabular}{lcccc}
    \toprule
    Backbone
    & \multicolumn{2}{c}{Base}
    & \multicolumn{2}{c}{SNAP} \\
    \cmidrule(lr){2-3} \cmidrule(lr){4-5}
    & Top-1 $\uparrow$ & Top-5 $\uparrow$
    & Top-1 $\uparrow$ & Top-5 $\uparrow$ \\
    \midrule

    DINOv3~\cite{simeoni2025dinov3}
    & 68.45 & 87.38
    & \textbf{68.81} & \textbf{87.41} \\

    \rowcolor[HTML]{EFEFEF}
    SAM~\cite{Kirillov_2023_ICCV}
    & 18.06 & 34.45
    & \textbf{21.46} & \textbf{38.28} \\

    MAE~\cite{he2022mae}
    & \textbf{30.78} & \textbf{49.59}
    & 28.64 & 46.98 \\

    \bottomrule
  \end{tabular}
\end{table}

\begin{table}[!tbp]
  \vspace{-15pt}
  \centering
  \small
  \setlength{\tabcolsep}{7.0pt}
  \caption{
    \textbf{ADE20K semantic segmentation with frozen representations.}
    Linear-probe performance is reported in terms of mean
    Intersection-over-Union (mIoU) and pixel accuracy (\%).
    Best results are highlighted in \textbf{bold}.
  }
  \label{tab:teacher_ade20k}

  \begin{tabular}{lcc}
    \toprule
    Representation & mIoU $\uparrow$ & Pixel Acc. $\uparrow$ \\
    \midrule

    DINOv3~\cite{simeoni2025dinov3}
    & \textbf{51.92} & \textbf{87.27} \\

    \rowcolor[HTML]{EFEFEF}
    SNAP
    & 49.87 & 84.31 \\

    \bottomrule
  \end{tabular}
\end{table}

\subsection{Comparison with V-JEPA}
\label{sec:appendix_vjepa}

SNAP and V-JEPA share the use of feature-space prediction rather than
direct pixel reconstruction, but they differ fundamentally in the type of representation they are designed to learn. V-JEPA~\cite{bardes2024vjepa} is a video representation learner that predicts latent features across video observations, while V-JEPA 2.1~\cite{murlabadia2026vjepa21unlockingdense} extends this paradigm toward dense video features. In contrast, SNAP is explicitly a multi-view encoder: it learns a shared scene representation from multiple posed images and trains that representation to predict features at arbitrary target viewpoints. Thus, the comparison separates the benefit of feature-space prediction from the additional geometric constraint introduced by explicit multi-view representation learning.

\paragraph{Hypersim, RealEstate10K and DL3DV.}
We evaluate a V-JEPA 2.1 ViT-L model using the same frozen-feature
point-correspondence protocol as SNAP, i.e., global soft-argmax matching with a low temperature. As shown in Table~\ref{tab:vjepa_comparison}, SNAP outperforms both V-JEPA 2.1-L and \dinovthree on RealEstate10K, DL3DV and Hypersim at every threshold, with the largest margins at tight thresholds.

On Hypersim, SNAP achieves 55.35, 77.90 and 88.35 PCK at thresholds of 0.1, 0.2 and 0.3, respectively, compared with 21.45, 50.21 and 72.44 for V-JEPA 2.1-L. \dinovthree obtains 29.29, 63.67 and 84.22. On RealEstate10K, SNAP achieves 78.21, 91.32 and 95.62, compared with 35.85, 67.05 and 85.05 for V-JEPA 2.1-L and 45.83, 80.13 and 93.71 for \dinovthree. DL3DV shows the same pattern (PCK@0.1: 58.74 for SNAP vs.\ 24.55 and 28.00). The advantage narrows at PCK@0.3, where all methods approach saturation; on RealEstate10K the SNAP--\dinovthree gap there is under 2 points.

\begin{table}[!tbp]
  \centering
  \footnotesize
  \setlength{\tabcolsep}{3pt}
  \setlength{\aboverulesep}{0pt}
  \setlength{\belowrulesep}{0pt}
  \renewcommand{\arraystretch}{1.3}
  \caption{
    \textbf{Point correspondence compared with V-JEPA 2.1.}
    PCK (\%) at thresholds $0.1$, $0.2$ and $0.3$ for
    DINOv3~\cite{simeoni2025dinov3} (ViT-B), SNAP (ViT-B), and
    V-JEPA 2.1-L~\cite{murlabadia2026vjepa21unlockingdense}, using global
    soft-argmax matching ($\tau=0.02$). Values are mean $\pm$ CI.
    Hypersim is held out from SNAP's training. Best results are
    highlighted in \textbf{bold}.
  }
  \label{tab:vjepa_comparison}
  \resizebox{\textwidth}{!}{
  \begin{tabular}{l ccc @{\hspace{10pt}} ccc @{\hspace{10pt}} ccc}
    \toprule
    & \multicolumn{3}{c}{RealEstate10K}
      & \multicolumn{3}{c}{DL3DV}
      & \multicolumn{3}{c}{Hypersim} \\
    \cmidrule(lr){2-4} \cmidrule(lr){5-7} \cmidrule(lr){8-10}
    Method
    & @0.1 & @0.2 & @0.3
    & @0.1 & @0.2 & @0.3
    & @0.1 & @0.2 & @0.3 \\
    \midrule

    DINOv3~\cite{simeoni2025dinov3}
    & 45.83\ci{0.80} & 80.13\ci{0.74} & 93.71\ci{0.47}
    & 28.00\ci{0.65} & 62.53\ci{0.87} & 84.45\ci{0.67}
    & 29.29\ci{0.73} & 63.67\ci{0.95} & 84.22\ci{0.79} \\

    \rowcolor[HTML]{EFEFEF}[\tabcolsep]
    V-JEPA 2.1-L~\cite{murlabadia2026vjepa21unlockingdense}
    & 35.85\ci{0.66} & 67.05\ci{0.77} & 85.05\ci{0.62}
    & 24.55\ci{0.55} & 54.94\ci{0.76} & 77.27\ci{0.67}
    & 21.45\ci{0.52} & 50.21\ci{0.80} & 72.44\ci{0.79} \\

    SNAP
    & \textbf{78.21}\ci{1.06} & \textbf{91.32}\ci{0.78} & \textbf{95.62}\ci{0.59}
    & \textbf{58.74}\ci{1.14} & \textbf{80.27}\ci{1.02} & \textbf{89.61}\ci{0.81}
    & \textbf{55.35}\ci{0.98} & \textbf{77.90}\ci{0.95} & \textbf{88.35}\ci{0.80} \\

    \bottomrule
  \end{tabular}%
  }
\end{table}

The consistent advantage across all three benchmarks indicates that
feature-space prediction alone is insufficient to obtain SNAP's level of
geometric transfer. The explicit multi-view constraint requires the
representation to remain useful across viewpoint changes, rather than only
across temporally related video observations.

Overall, the comparison highlights a fundamental distinction between
video/image representation learning and explicit multi-view representation
learning. Although both V-JEPA~\cite{bardes2024vjepa} and SNAP operate in feature space, their training constraints are different. V-JEPA learns representations through latent prediction between video observations, whereas SNAP learns a shared scene representation from multiple posed images and requires that representation to support feature prediction across viewpoints. The stronger correspondence performance of SNAP across Hypersim, RealEstate10K, and DL3DV indicates that this explicit multi-view geometric constraint provides a distinct advantage beyond feature-space prediction alone.

\section{Inference Speed}
\label{sec:appendix_inference}

We evaluate the inference efficiency of SNAP against Muskie~\cite{li2025muskiemultiviewmaskedimage} and VGGT~\cite{wang2025vggt} to characterize the computational overhead of our architecture. As shown in Table~\ref{tab:encoder_efficiency}, SNAP demonstrates superior scalability and memory efficiency as the number of input views ($V$) increases.

\paragraph{Memory and Parameter Efficiency.} 
SNAP is significantly more memory-efficient than both baselines. At a high-resolution regime of $V=100$, SNAP requires only \textbf{1.03 GB} of peak GPU memory, representing a \textbf{40.5\% reduction} compared to Muskie (1.73 GB) and a \textbf{69.8\% reduction} compared to VGGT (3.41 GB). This efficiency is underpinned by SNAP's compact parameterization; with only \textbf{233.38M} parameters including the \dinovthree backbone, SNAP is approximately $1.3\times$ smaller than Muskie and $3.9\times$ smaller than the geometry-grounded VGGT. 

\paragraph{Temporal Performance.} 
While SNAP consistently outperforms VGGT, achieving a \textbf{2.9$\times$} speedup at $V=100$, it is roughly \textbf{36\% slower} than Muskie in the high-view regime (0.245s vs. 0.180s). We attribute this latency gap to Muskie's architecture, which utilizes fewer multi-view self-attention blocks compared to our design. However, for practical deployment, SNAP offers a more accessible memory profile: at $V=100$, SNAP requires only \textbf{1.03 GB} of VRAM, whereas Muskie consumes $67.9\%$ more memory. For practitioners working with limited GPU memory or ultra-dense view counts, SNAP provides a viable alternative that prioritizes a smaller memory footprint over raw inference throughput. Notably, SNAP provides a different tradeoff of performance, speed, and memory. 

\begin{table}[!t]
\vspace{-15pt}
\caption{
\textbf{Encoder forward-pass efficiency vs.\ number of input views.}
For each encoder we report forward-pass time (milliseconds), peak GPU memory, and number of parameters.
All encoders are benchmarked under identical conditions: bfloat16, single forward call per measurement, and we utilized an NVIDIA RTX PRO 6000 Blackwell GPU.
}

\centering

\resizebox{\linewidth}{!}{%
\begin{tabular}{llcccccccc}
\toprule
Method & Metric & $V{=}1$ & $V{=}2$ & $V{=}4$ & $V{=}8$ & $V{=}10$ & $V{=}20$ & $V{=}50$ & $V{=}100$ \\

\midrule
Muskie~\cite{li2025muskiemultiviewmaskedimage}
  & Time (ms) $\downarrow$   & \textbf{12} & \textbf{13} & \textbf{13} & \textbf{16} & \textbf{16} & \textbf{25} & \textbf{67} & \textbf{180} \\
  & Mem (GB) $\downarrow$   & 0.59  & 0.61  & 0.64  & 0.67  & 0.70  & 0.81  & 1.16  & 1.73  \\
  & Params (M) $\downarrow$ & 303.10 & 303.10 & 303.10 & 303.10 & 303.10 & 303.10 & 303.10 & 303.10 \\
  
\midrule
\rowcolor[HTML]{EFEFEF}
VGGT~\cite{wang2025vggt}
  & Time (ms) $\downarrow$   & 60 & 55 & 41 & 71 & 89 & 123 & 294 & 712 \\
  \rowcolor[HTML]{EFEFEF}
  & Mem (GB) $\downarrow$   & 2.18  & 2.20  & 2.22  & 2.27  & 2.29  & 2.42  & 2.79  & 3.41  \\
  \rowcolor[HTML]{EFEFEF}
  & Params (M) $\downarrow$ & 909.11 & 909.11 & 909.11 & 909.11 & 909.11 & 909.11 & 909.11 & 909.11 \\

\midrule
SNAP
    & Time (ms) $\downarrow$  & 20 & 23 & 23 & 22 & 22 & 30 & 85 & 245 \\
  & Mem (GB) $\downarrow$   & \textbf{0.45}  & \textbf{0.46}  & \textbf{0.49}  & \textbf{0.49}  & \textbf{0.51}  & \textbf{0.57}  & \textbf{0.74}  & \textbf{1.03}  \\
  & Params (M) $\downarrow$ & 233.38 & 233.38 & 233.38 & 233.38 & 233.38 & 233.38 & 233.38 & 233.38 \\

\bottomrule
\end{tabular}
}
\label{tab:encoder_efficiency}
\end{table}

\section{Qualitative Depth Estimation Results}
\label{sec:appendix_qualitative_depth}

\begin{figure}[!t]
  \centering
  \includegraphics[width=0.90\linewidth]{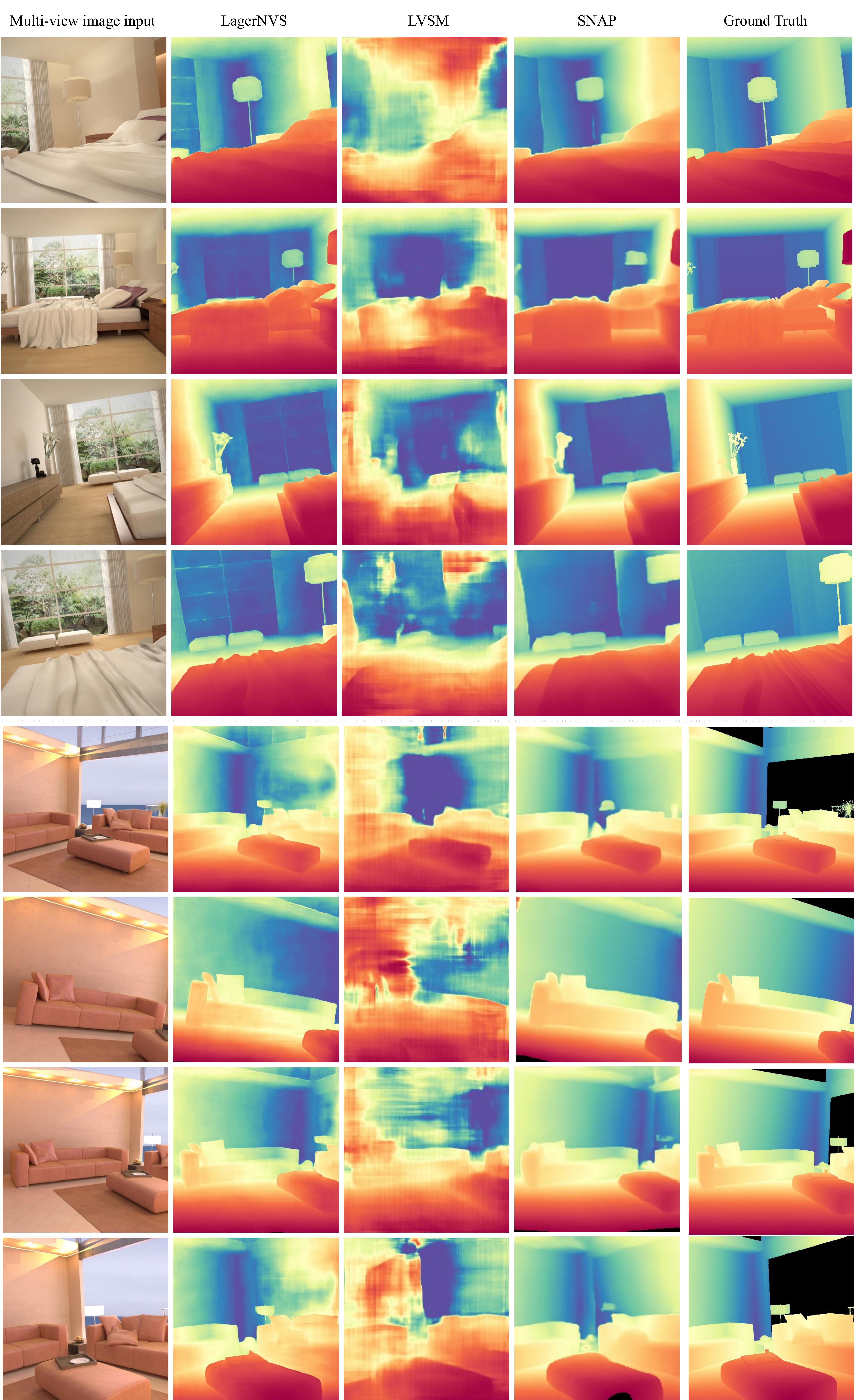}
  \caption{
    \textbf{Qualitative Depth Evaluation}: Hypersim relative depth predictions from a sparse collection of four input images using a frozen backbone with trained DPT layers. SNAP features yield depth predictions comparable to LagerNVS, despite using a lower decoding resolution (256px vs.\ 518px for LagerNVS). The higher resolution of LagerNVS likely contributes to its stronger predictions on thin structures.
  }
  \label{fig:qualitative_depth}
\end{figure}

We provide qualitative relative depth estimation results from our trained probe (see Figure~\ref{fig:qualitative_depth}). Given a sparse collection of four input images, we use a frozen backbone with trained DPT layers to predict relative depth. Overall, SNAP produces coherent depth predictions that are comparable to those of LagerNVS (\emph{geometry pretrained}). In contrast, LVSM exhibits noisier predictions and less consistent depth structure across the examples. 

The comparison with LagerNVS is not resolution-matched: LVSM and SNAP are decoded at 256px, whereas LagerNVS is decoded at 518px. The higher decoding resolution of LagerNVS likely contributes to its stronger depth estimation.

\section{Landscape Plots}
\label{sec:appendix_landscape}
\begin{figure}[t]
  \centering
  \includegraphics[width=\linewidth]{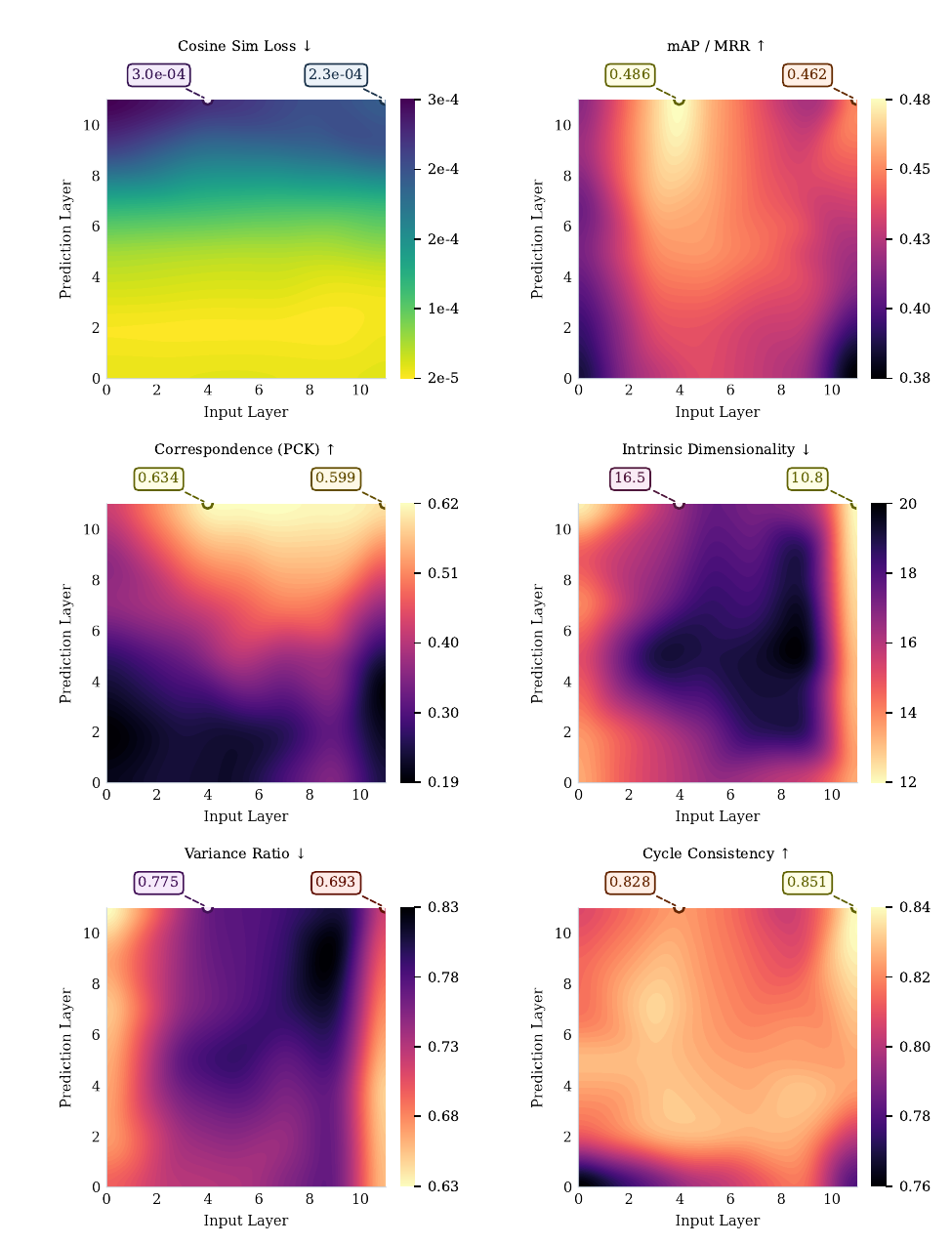}
  \caption{
    \textbf{Extended Input vs. Prediction Layer Landscape.} We observe a fundamental trade-off depending on the chosen input layer (yellow indicates superior performance across all metrics). While mid-layer inputs (e.g., Layer 4) optimize local discriminability (PCK, mAP/MRR), they create a distinct performance valley in holistic metrics. To maximize cycle consistency and scene clustering (Variance Ratio), initializing and predicting with the deepest layers (Layer 11) proves to demonstrate a robust optimal operating point.
  }
  \label{fig:landscape_expanded}
\end{figure}

We expand the sweep methodology from Section \ref{sec:axis2} to evaluate a broader set of metrics: mAP/MRR (reflecting bag-of-words correspondence retrieval), cycle consistency, and variance ratio (the ratio of within-scene to across-scene variance, indicating how well features cluster by scene) (Figure~\ref{fig:landscape_expanded}). Interestingly, we observe a distinct trade-off. While pairing mid-layer inputs with last-layer prediction targets yields higher raw correspondence, this configuration results in lower cycle consistency compared to initializing with the last layer of DINO. This performance valley for mid-layer inputs is consistently prominent across mAP/MRR, cycle consistency, and the variance ratio. Ultimately, this highlights that optimizing for raw local correspondence does not necessarily yield the most structurally robust, 3D-grounded representations, justifying our use of deeper semantic layers to enforce global multi-view consistency.

\section{Qualitative Feature Visualizations}
\begin{figure}[t]
    \centering

    \includegraphics[width=\linewidth]{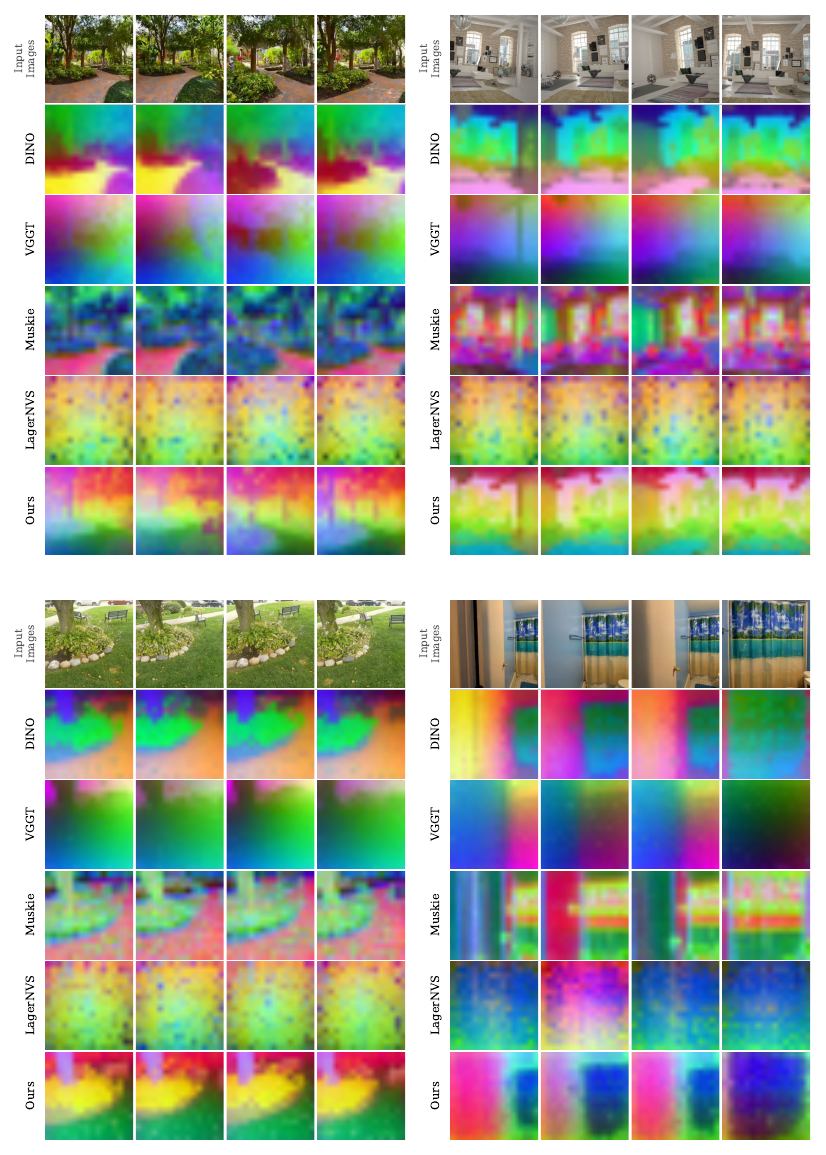}
    \caption{
    \textbf{Feature map visualizations across vision encoders.} High-dimensional latents are projected to RGB space using PCA to visualize spatial abstraction. While the 2D teacher (DINO) produces flat semantic regions, \method introduces clear spatial gradients that track underlying 3D geometry and depth, all while preserving fine-grained, discriminative object boundaries. Baselines struggle with this geometric abstraction: Muskie heavily encodes raw 2D appearance, and LagerNVS exhibits severe spatial grid artifacts.
    }
    \label{fig:pca_qualitative}
\end{figure}

To qualitatively assess what these models learn, we visualize their high-dimensional feature maps projected into RGB space. As illustrated in Figure \ref{fig:pca_qualitative}, standard 2D priors like DINO group scenes into flat, semantic regions but lack fine-grained 3D geometric structure. In contrast, \method clearly learns a representation distinct from its \dinovthree teacher: it introduces smooth, continuous spatial gradients across physical surfaces (e.g., ground planes), indicating an emergent understanding of 3D geometry while preserving fine-grained, discriminative object boundaries. Alternative multi-view baselines fail to achieve this geometric abstraction. Muskie heavily encodes raw appearance and texture rather than underlying structure, resulting in representations that simply mimic the RGB input. Meanwhile, LagerNVS suffers from severe, high-frequency grid artifacts, disrupting the continuous spatial coherence required for robust multi-view reasoning along with feature anisotropy.


\clearpage

\end{document}